\documentclass[11pt]{article}

\usepackage[margin=1in]{geometry}
\usepackage{times}
\usepackage[hyphens]{url}
\usepackage{graphicx}
\usepackage{natbib}
\usepackage{caption}
\usepackage{subcaption}
\usepackage{algorithm}
\usepackage{algorithmic}
\usepackage{amsmath}
\usepackage{amsthm}
\usepackage{multirow}
\usepackage{color}
\usepackage{framed}
\usepackage{booktabs}
\usepackage{amsfonts}
\usepackage{amssymb}
\usepackage{threeparttable}
\usepackage[table]{xcolor}
\usepackage{hyperref}
\usepackage{subcaption}

\DeclareMathAlphabet{\bbold}{U}{bbold}{m}{n}

\theoremstyle{plain}
\newtheorem{theorem}{Theorem}

\newtheorem{proposition}{Proposition}
\newtheorem{lemma}{Lemma}

\newtheorem{definition}{Definition}

\theoremstyle{remark}
\newtheorem{remark}{Remark}

\renewcommand{\Pr}{\mathbb{P}}

\newcommand{\bomega}{\boldsymbol{\omega}}
\newcommand{\bx}{\mathbf{x}}
\newcommand{\bSigma}{\boldsymbol{\Sigma}}

\title{Partial Fairness Awareness: Belief-Guided Strategic Mechanism for Strategic Agents}

\author{
    Xinpeng Lv$^{1}$,
    Chunyuan Zheng$^{2}$,
    Yunxin Mao$^{1}$,
    Renzhe Xu$^{3}$,
    Hao Zou$^{4}$,
    Shanzhi Gu$^{1}$,
    Liyang Xu$^{1}$,\\
    Huan Chen$^{1}$,
    Yuanlong Chen$^{5}$,
    Wenjing Yang$^{1}$,
    Haotian Wang$^{1\dagger}$
    \\[6pt]
    $^1$National University of Defense Technology, Changsha, China\\
    $^2$Peking University, Beijing, China\\
    $^3$Shanghai University of Finance and Economics, Shanghai, China\\
    $^4$ZGC Laboratory, Beijing, China\\
    $^5$Faculty of Computing, Harbin Institute of Technology, Harbin, China\\
    {\small \texttt{\{lvxinpeng, maoyunxin, wanghaotian13\}@nudt.edu.cn}}\\
    {\small  $^\dagger$Corresponding author.}
}

\date{}

\begin{document}

\maketitle

\begin{abstract}
Strategic machine learning investigates scenarios where agents manipulate their features to receive favorable decisions from predictive models. To address fairness concerns intrinsic to strategic classification, recent work has introduced group-specific fairness constraints. However, current fairness-aware approaches face a fundamental dilemma in the issue of fairness exposure: making these constraints public enables strategic manipulation and can lead to fairness reversal, while keeping them hidden may reduce social welfare and discourage genuine improvement.
To fill this gap, we subsequently propose the problem of \textit{\textbf{P}artial \textbf{F}airness \textbf{A}wareness} (PFA), as our theoretical analysis informs that such a dilemma can be mitigated by releasing the candidate set of fairness constraints and concealing the grounding constraint.
To be specific, we introduce a \textbf{belief-guided strategic mechanism}, wherein agents iteratively interact with the decision system and maintain a belief distribution over the candidate set of fairness constraints. This belief-guided process enables agents, through iterative interaction and feedback, to update their belief distribution over the candidate set, thereby gradually aligning their belief with the grounding fairness constraint employed by the system.
Extensive experiments on real-world and synthetic datasets demonstrate that PFA achieves lower group fairness gaps, higher acceptance of truly qualified individuals, and more stable outcomes compared to fully public or private fairness regimes.
\end{abstract}

\section{Introduction}

Machine learning models are increasingly deployed in decision-making domains such as hiring~\cite{sanchez2020does}, credit scoring~\cite{jagtiani2019roles}, and college admissions~\cite{kuvcak2018machine}. In these scenarios, individuals (agents) often engage in strategic manipulation of their features to obtain favorable outcomes. As noted by Goodhart's Law~\cite{Strathern_1997}, ``\textit{When a measure becomes a target, it ceases to be a good measure},'' such gaming behaviors can undermine the reliability of decision models.
For example, a loan applicant might temporarily inflate their reported income to appear more creditworthy. To ensure robustness towards the strategic manipulations, the strategic classification~(SC) framework~\cite{hardt2016strategic} has been developed to model Stackelberg-style interaction between model designers and strategic agents~\cite{pmlr-v139-ghalme21a,singh2024optimal,chen2020learning}, aiming to maintain predictive accuracy under adversarial conditions.

Despite the remarkable progress of the SC framework in robustness, practical deployment also necessitates careful consideration of fairness for sensitive attributes or historically disadvantaged groups. In particular, a model resistant to gaming may still systematically disadvantage certain groups, i.e., exhibiting \textit{unfairness}. For example, even if a hiring platform exhibits robustness to manipulation, historical biases embedded in the data might lead to unfair judgment towards female applicants, reflecting and reinforcing social inequities~\cite{zemel2013learning,zhang2022fairness}. To address the \textit{fairness concern in strategic-robust decision models}, a growing body of work has incorporated group fairness constraints, such as demographic parity~\cite{zemel2013learning}, equality of opportunity~\cite{roemer2015equality}, and predictive parity~\cite{dieterich2016compas}, into strategic classification, typically through fairness-aware decision rules adapted for different groups~\cite{zhang2022fairness,shimao2025strategic}.

\begin{figure*}[t]
    \centering
    \begin{subfigure}{\textwidth}
    \centering
    \includegraphics[width=0.6\textwidth]{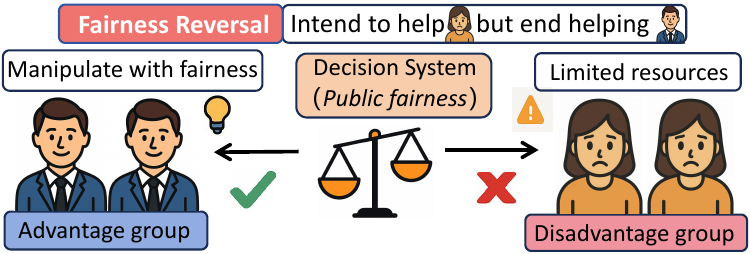}
    \caption{Public Fairness}
    \label{fig1a}
\end{subfigure}
    \begin{subfigure}{\textwidth}
    \centering
    \includegraphics[width=0.6\textwidth]{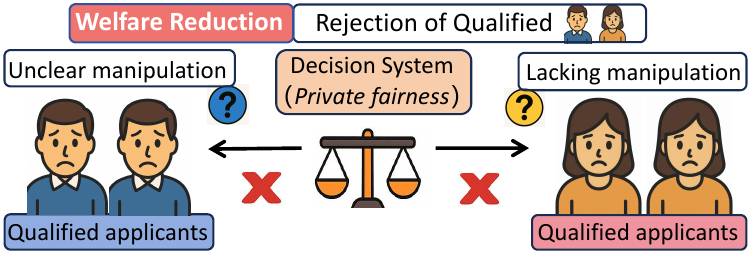}
    \caption{Private Fairness}
    \label{fig1b}
\end{subfigure}
    \begin{subfigure}{\textwidth}
    \centering
    \includegraphics[width=0.6\textwidth]{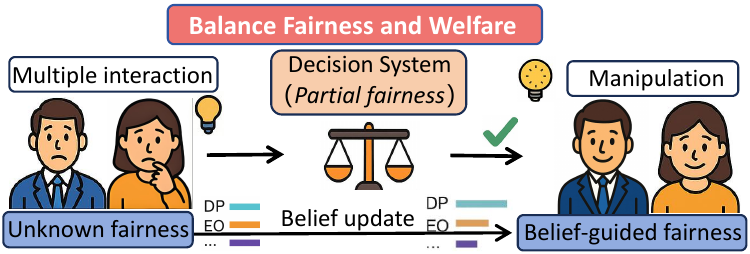}
    \caption{Partial Fairness with Belief-guided}
    \label{fig1c}
\end{subfigure}
    \caption{An illustration for the fairness challenge in strategic classification. (a) Public fairness allows strategic behavior from advantaged agents, leading to fairness reversal~\textit{(left)}. (b) Private fairness avoids manipulation but rejects qualified individuals, causing social welfare loss~\textit{(center)}. (c) Our belief-guided strategic mechanism enables sequential learning for agents with partial fairness, improving group welfare and preventing fairness reversal~\textit{(right)}.}
    \label{fig1}
\end{figure*}

However, most concurrent efforts focus on designing more effective fairness mechanisms which is fully public to agents, while overlooking a more fundamental consideration: \textit{whether and how fairness constraints should be exposed to agents}.
In practice, different exposure of the fairness constraints results in divergent subsequent fairness mechanism design, leading to different algorithmic choices and social welfare variation.
On the one hand, in the case of fully public constraints, previous theories already point out the unexpected phenomenon of \emph{fairness reversal}~\cite{fairnessreverse}~(public fairness in Figure~\ref{fig1a}), where advantaged individuals manipulate based on the exposed fairness constraints. Hence, the reverse effect of fairness occurs, as the intended benefits for disadvantaged groups are conversely further undermined.
On the other hand, when the fairness constraints are fully concealed~(private fairness in Figure~\ref{fig1b}), our theoretical analysis proves that the social welfare, i.e., the acceptance rate of the qualified individuals, will decrease. Subsequently, reinforcing concealed fairness constraints leads to less incentivized model design, violating \textit{the principle of incentive alignment}~\cite{haghtalab2020maximizing}. Combining the two sides, concurrent approaches on strategic fairness reveal a fundamental {\bf dilemma}:
\begin{framed}
\textit{Fully public constraints can lead to fairness reversals and exacerbate inequality, while fully private constraints risk reducing overall social welfare.}
\end{framed}

To address this dilemma, we introduce the novel problem \textit{Partial Fairness Awareness} (PFA), a trade-off that balances the extremes of full transparency and privacy in fairness-aware strategic classification. In PFA, only a candidate set of fairness constraints is released to agents while the grounding constraint remains concealed.
To fill this gap, we design a \textbf{belief-guided strategic mechanism}, where agents sequentially interact with the decision system and maintain a belief distribution over the candidate fairness constraints. At each round of interaction, the agent selects its optimal manipulation strategy based on its current belief, and updates the belief distribution with the feedback from the system~(as shown in Figure~\ref{fig1c}). As interactions proceed, this sequential learning process enables the agents to gradually infer and align with the actual (grounding) fairness constraint employed by the system. Our theory and empirical results demonstrate that this belief-guided mechanism effectively mitigates fairness reversal and improves social group welfare.

\textbf{Our main contributions are summarized as follows:}
\begin{itemize}
\item We conduct a systematic theoretical analysis of the fundamental dilemma between public and private disclosure of fairness constraints in strategic classification. To address the limitations of these two extremes, we formulate a novel problem of \textbf{Partial Fairness Awareness (PFA)}.
\item To tackle the PFA problem, we design a \textbf{belief-guided strategic mechanism}, formally modeled via Bayesian inference. We rigorously prove that, under this mechanism, agents' beliefs converge to the true fairness constraint employed by the system, thereby improving both group fairness and overall social welfare.
\item Extensive experiments on both synthetic and real-world datasets demonstrate that the PFA mechanism significantly outperforms conventional approaches based on fully public or fully private fairness constraints, with respect to fairness, social welfare, and predictive accuracy.
\end{itemize}

\section{Related Work}
\label{relatedwork}

\subsection{Strategic Machine Learning}
Strategic classification~\cite{hardt2016strategic} studies settings where individuals manipulate their features to influence model outcomes~\cite{dong2017strategicclassificationrevealedpreferences,shavit2020causal,chen2020learning,NEURIPS2021_f1404c26,zrnic2021leads,tsirtsis2024optimal,lv2026beyond,lv2026breaking,lv2026tabular}. Several works address challenges arising from unknown manipulations or limited agent information~\cite{shao2024strategic,pmlr-v139-ghalme21a,11097075}.
Recent studies incorporate causal reasoning into strategic learning~\cite{miller2020strategic,chen2023learning,horowitz2023causal,vo2024causal,efthymiou2025incentivizing,chang2024s,10.1007/978-3-032-05962-8_25,wang2022estimating,wang2023treatment}, distinguishing between manipulable and genuinely improvable features. This line of work emphasizes how strategic responses may reflect or distort true underlying qualifications.
Relatedly, performative prediction~\cite{perdomo2020performative,hardt2022performative,hardt2023performative,mendler2022anticipating,mofakhami2023performative} analyzes how predictive models influence the data distribution over time through repeated deployment.
A complementary direction focuses on promoting social welfare~\cite{haghtalab2020maximizing,estornell2023incentivizing,xie2024non}, designing mechanisms that align agent incentives with collective benefit.

\subsection{Fairness-aware Strategic Classification}

Recent research has examined the complex interplay between fairness and strategic behavior in machine learning. Several studies highlight that unequal manipulation costs can exacerbate disparities even in the presence of fairness constraints~\cite{hu2019disparate,milli2019social}.
To mitigate these issues, various methods have been proposed, such as optimizing classifiers to reduce manipulation costs for disadvantaged groups~\cite{keswani2023addressing} or employing minimax group fairness frameworks~\cite{diana2024minimax,zhengchunyuan}. Other work evaluates fairness through agents' equilibrium behaviors~\cite{shimao2025strategic,yang2024your,wang2025simprof} and explores how incentive structures can influence manipulation~\cite{zhang2022fairness}. Moreover, group fairness constraints may unintentionally result in fairness reversal when agents strategically modify their features~\cite{fairnessreverse}.
Recent efforts also focus on constructing fairness-aware models that anticipate manipulation and promote genuine improvement~\cite{alhanouti2025anticipating}.

\section{Preliminary}
\label{pre}

We present the essential background on strategic machine learning and causal decorrelation methods. Throughout this paper, we denote random variables by uppercase letters~(e.g., $X$ and $Y$) and their realizations by lowercase letters~(e.g., $x$ and $y$). Bold symbols (e.g., $\mathbf{x}$ and $\mathbf{X}$) are used for vectors or matrices.

\subsection{Strategic Classification}
The strategic classification problem is modeled as a Stackelberg game~\cite{li2017review}, where a \textbf{decision maker} defines a classification function \(f: \mathbb{R}^d \to \{0,1\}\), and \textbf{decision subjects} (agents) strategically manipulate their features from \(\mathbf{x}\) to \(\mathbf{x}'\) at a cost \(c: \mathcal{X} \times \mathcal{X} \to \mathbb{R}_{\geq 0}\)~\cite{hardt2016strategic,miller2020strategic}.

The optimal manipulated feature \(\mathbf{x}'\) is determined by the best-response function \(b(\mathbf{x})\):

\begin{definition}[Strategic Manipulation]
    \begin{equation}
    \mathbf{x}' = b(\mathbf{x}) = \arg \max_{x \in \mathcal{D}} U(\mathbf{x},\mathbf{x}'),
    \label{bestreponse}
\end{equation}
where $\mathcal{D}$ denotes the distribution of the agents' features. Specifically, the utility function $U(\mathbf{x}, \mathbf{x}')$ is given by
\begin{equation}
    U(\mathbf{x},\mathbf{x}')= f(\mathbf{x}') - \lambda c(\mathbf{x}, \mathbf{x}'),
    \label{utilityfunc}
\end{equation}
where \(f(\mathbf{x}') \in \{0,1\}\) represents the classification outcome under the modified features and \(c(\mathbf{x}, \mathbf{x}')\) is the cost associated with modification. And \(\lambda > 0\) is a trade-off parameter that balances the classification benefit against the associated cost.
\end{definition}

To avoid the impact caused by strategic behaviors, the decision maker optimizes decision rules \(f'\) to maximize expected classification accuracy:
\begin{definition}[Decision Optimization]
    \begin{equation}
    f' \;\in\; \arg\max_{f \in \mathcal{F}} \;\mathbb{E}_{(\mathbf{x},y)\sim \mathcal{D}}\bigl[\bbold{1}\bigl(f(b(\mathbf{x})) = y\bigr)\bigr],
\end{equation}
where \(\mathcal{F}\) is the set of feasible decision rules, \(\mathbf{x}\) is the original feature vector in distribution \(\mathcal{D}\), and \(y\) is the agent's label.
\end{definition}

\subsection{Fairness-aware Strategic Classification}
In practical applications, strategic classification tasks are often required to satisfy group fairness constraints to ensure equitable outcomes for different demographic groups by imposing fairness constraints, such as demographic parity, on the classifier's decisions:
\begin{definition}[Fairness-aware SC]
    Given fairness metric $\mathcal{C}$ (e.g., demographic parity gap) with group $g$, the strategic classifier $f$ is optimized as:
    \begin{equation}
    \begin{aligned}
        &\max_{f \in \mathcal{F}}~\mathbb{E}_{(x, y; g) \sim D}\left[\mathbb{I}(f(b(\mathbf{x}; g)) = y)\right], \\
    &\quad \text{subject to}\quad \mathcal{C}(f; D, G) \leq \delta,
    \end{aligned}
    \end{equation}
    where $\mathcal{C}(f; D, G)$ quantifies the group fairness gap and $\delta$ controls the allowable disparity. Moreover, $b(\mathbf{x}; g)$ denotes group-dependent strategic manipulation (best response) for agents in group $g$. This dependence on $g$ arises because fairness constraints may affect groups differently, leading to distinct strategic behaviors.
\end{definition}

\subsection{Social Welfare in Strategic Classification}

While traditional SC models emphasize robustness against manipulation, they often overlook the possibility that agents may engage in \emph{genuine improvement}, that is, modifying their features in a way that reflects a true enhancement of their underlying qualification or ability~\cite{horowitz2023causal,chen2023learning}.

In many real-world settings, such improvements are not only possible but desirable, e.g., a job applicant might complete a relevant training program to enhance employability. As a result, \textit{genuinely improved} individuals may be misclassified and rejected, thereby discouraging effort and undermining long-term social welfare. More formally, we present the definition of \textit{social welfare} to measure the number of truly qualified individuals who are accepted by the classifier after strategic modification~\cite{haghtalab2020maximizing,estornell2023incentivizing}.

\begin{definition}[Social Welfare]
\label{def:welfare}
The \textit{social welfare} \( W \) is defined as
\begin{equation}
W = \sum_{i} \mathbb{I}(y_i \geq y_{\text{th}}) \cdot \mathbb{I}(s(x_i') \geq \theta_{g_i}),
\label{welfare}
\end{equation}
where \( \mathbb{I}(\cdot) \) is the indicator function, \( x_i' \) is the strategically modified feature vector, and \( \theta_{g_i} \) is the decision threshold for group \( g_i \). A higher value of \( W \) reflects better alignment between the classifier's decisions and agents' true improvement, and thus greater societal benefit.
\end{definition}

\begin{figure}[t]
    \centering
    \includegraphics[width=0.8\linewidth]{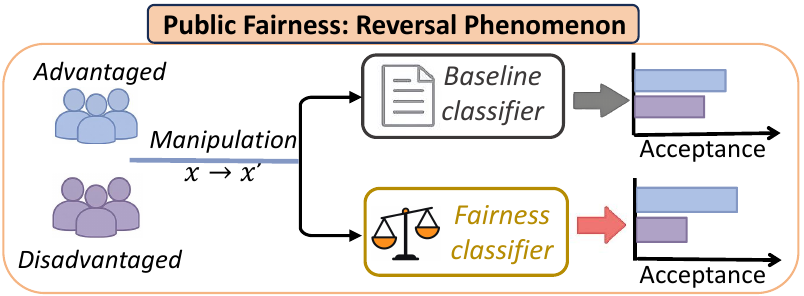}
    \caption{Illustration of the reversal phenomenon with public fairness in strategic classification.}
    \label{figre}
\end{figure}

\section{Dilemma of Strategic Fairness}

\subsection{Public Fairness: Reversal Phenomenon}

On the one hand, common SC approaches always assume the public accessibility of the fairness constraints. However, existing theories already inform that, \textit{the advantaged group will utilize the exposure of fairness constraints to further manipulations}, as shown in Figure~\ref{figre}, leading to exacerbated inequality~(i.e., compounded unfairness):

\begin{lemma}[Fairness Reversal~\cite{fairnessreverse}]
Let $f^\dagger$ be a classifier trained to satisfy a group fairness constraint (e.g., demographic parity), and let $f^\sim$ be a baseline classifier without a fairness constraint. In the context of strategic manipulation, the reversal phenomenon occurs, defined as:
\begin{equation}
\Delta_{\text{fair}}(f^\dagger, \mathcal{D}) > \Delta_{\text{fair}}(f^\sim, \mathcal{D}),
\end{equation}
where $\Delta_{\text{fair}}(\cdot, \mathcal{D})$ denotes the fairness gap~(e.g., difference in acceptance rates across groups) measured after strategic manipulation. That is, $f^\dagger$, though trained to be fair, exhibits a larger fairness gap than $f^\sim$ under agent manipulation.
\label{lemma:reversal}
\end{lemma}

\begin{figure}[t]
    \centering
    \includegraphics[width=0.8\linewidth]{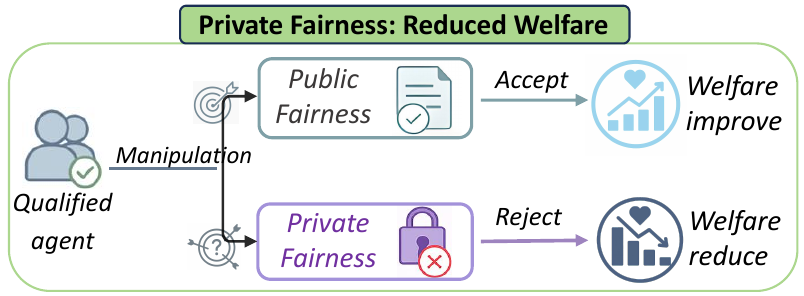}
    \caption{Illustration of reduced welfare: Private fairness constraints result in the rejection of qualified agents.}
    \label{figwe}
\end{figure}

\subsection{Private Fairness: Reduced Welfare}
\label{sec:theory}

On the other hand, a natural question falls in that \textit{what if when the fairness constraints are fully private}, without any exposure to the agents.
To address this, we further conduct an extensive theoretical analysis, informing that \textit{concealing fairness constraints leads to a reduction in social welfare} (as defined in Eq.~\eqref{welfare}), despite prior findings~\cite{bent2019algorithmic,mutlu2022contrastive} suggesting potential protection of group equality~(see detailed proofs in Appendix A):

\begin{theorem}[Welfare Reduction with Private Fairness]
Let \( \mathbb{E}[W_{\text{public}}] \) and \( \mathbb{E}[W_{\text{private}}] \) denote the expected social welfare under public and private fairness constraint settings, respectively. Then:
\begin{equation}
\mathbb{E}[W_{\text{private}}] < \mathbb{E}[W_{\text{public}}].
\end{equation}
\label{lemma:welfare}
\end{theorem}
\begin{remark}
    Such welfare loss illustrates a practical drawback of private fairness constraints, where even well-intentioned policies can unintentionally exclude qualified candidates and undermine the incentives for genuine improvement. As shown in Figure~\ref{figwe}, the expected social welfare under private fairness settings is always less than that under public settings.
\end{remark}

\noindent {\bf Additional Manipulation Costs.}
In addition, with private fairness constraints, agents cannot accurately target the system's acceptance criteria, resulting in unnecessary or repeated manipulation. Consequently, such misalignment leads to strictly higher manipulation costs compared to the public fairness setting~(see detailed proof in Appendix B).

\begin{proposition}
[Additional cost with private fairness]
Hiding group-specific fairness constraints leads to higher manipulation costs for both advantaged and disadvantaged groups than disclosing them.
\label{lemma3}
\end{proposition}

\begin{figure}[t]
    \centering
    \includegraphics[width=0.68\linewidth]{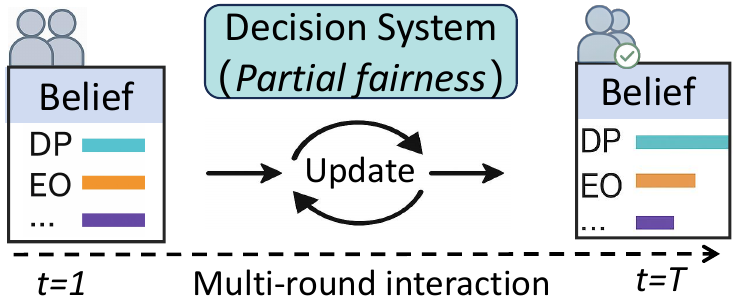}
    \caption{Illustration of belief-guided strategic mechanism: At each round, the agent updates its belief over the fairness mechanism set (e.g., DP, EO) based on feedback from the decision system, enabling gradual alignment with the true fairness constraint.}
    \label{fig2}
\end{figure}

\section{Partial Fairness Awareness Formulation and Belief-Guided Mechanism}

In this section, we formally define the Partial Fairness Awareness (PFA) problem in strategic classification, and present our belief-guided adaptation approach for learning and decision-making under this setting. We also analyze the impact of PFA and the belief-guided strategic mechanism on fairness-aware strategic classification.

\subsection{Partial Fairness Awareness in Strategic Classification}

To address the fairness dilemma in strategic classification, we naturally propose the viewpoint that \textit{the fairness constraints should be partially exposed to the agents}, termed as \emph{\textbf{P}artial \textbf{F}airness \textbf{A}wareness (PFA)}. To be specific, our PFA problem only lets agents access the set of possible fairness constraints~(i.e., the information set), while leaving the fact that which constraint is indeed adopted by the decision-maker unobserved. Therefore, agents have to infer their own belief distributions from interactions with the decision-maker in the \textit{sequential learning} process.

\begin{definition}[Fairness Mechanism Set]
    Let $\mathcal{M}$ denote the fairness mechanism set, i.e., the collection of all candidate fairness constraints the system may employ. The true, but unobserved, mechanism in effect is denoted by $\mathbf{m}^* \in \mathcal{M}$.
\end{definition}
\noindent For example, $\mathcal{M} = \{\texttt{None}, \texttt{DP}, \texttt{PP}, \texttt{EO}\}$ may represent the options of no constraint, demographic parity, and equalized odds, respectively, with $\mathbf{m}^* = \texttt{DP}$ indicating that demographic parity is adopted by the decision-maker.

\begin{definition}[Partial Fairness Awareness]
Given a fairness mechanism set $\mathcal{M}$ comprising all plausible group fairness constraints the system may employ, and the true but unknown mechanism $\mathbf{m}^* \in \mathcal{M}$, the agent's knowledge is limited to the set $\mathcal{M}$, not the identity of $\mathbf{m}^*$.
\end{definition}

Consequently, based on the fairness mechanism set, we then define the concept of \textit{belief distribution} over the fairness mechanism set, featuring the confidence of each agent in each candidate mechanism:

\begin{definition}[Belief Distribution]
    The \textit{belief distribution}~(probability distribution) is defined as $\mathbf{b} = \{ b^{(m)} \}_{m \in \mathcal{M}}$, where each $b^{(m)}$ represents the agent's belief in the fairness mechanism $m \in \mathcal{M}$.
\end{definition}

Notably, our sequential setup allows multi-round interactions between agents and the decision-maker, with the time step denoted as $t \in [T]$. Hence, in the initial time step, the agents will have an initial belief in a set of possible mechanisms~(e.g., $\mathbf{b}_0 = \{ b_0^{(m)} \}_{m \in \mathcal{M}}$), and gradually refine this belief~(e.g., $\mathbf{b}_t$) in each time step $t$ through repeated interactions and observed decision results via the strategic process of \textit{Belief-Guided Manipulation}:

\begin{definition}[Belief-guided Manipulation]
At each round $t$, the agent selects the belief-guided strategic manipulation with current belief $\mathbf{b}_t$, by maximizing expected utility under fairness uncertainty.
\end{definition}

This setup reflects our intuitions spanning over lots of realistic scenarios, where agents begin with no knowledge of the system's internal fairness mechanism and must learn over time through interaction~(as shown in Figure~\ref{fig2}).

\subsection{Belief-Guided Strategic Mechanism for PFA}

Within the concepts introduced above, we then detail the sequential policies of agents in each round, including \textit{how the belief of fairness constraint} is updated, and \textit{how agents strategically manipulate} based on the current belief.

\subsubsection{Belief Initialization.}
At each round $t$, the agent maintains a normalized belief distribution $\mathbf{b}_t = \{ b_t^{(m)} \}_{m \in \mathcal{M}}$, where $\sum_{m \in \mathcal{M}} b_t^{(m)} = 1$ and $b_t^{(m)} \in [0,1]$.
If no prior knowledge is assumed, the initial belief $\mathbf{b}_0$ is set uniformly over $\mathcal{M}$ (e.g., $b_{t=0}^{(m)} = 0.25$ for $|\mathcal{M}| = 4$).

\subsubsection{Belief-guided Strategic Manipulation.}
At the round $t$, given the current belief $\mathbf{b}_t$, the agent aims to select a feature modification~$\mathbf{x}' \in \mathcal{X}$ that maximizes their expected utility for getting favorable results. Formally, the agent solves the following optimization problem:
\begin{equation}
\begin{aligned}
    \mathbf{x}' = \arg\max_{\mathbf{x}' \in \mathcal{X}} \big( \mathbb{E}_{m \sim \mathbf{b}_t} \Pr_{f^{(m)}}&(f^{(m)}(\mathbf{x}') = 1) - \lambda\!\cdot\!c(\mathbf{x}, \mathbf{x}') \big),
    \label{belief-guided-m}
\end{aligned}
\end{equation}
where $\Pr_{f^{(m)}}(f^{(m)}(\mathbf{x}') = 1)$ denotes the estimated probability of being accepted by the classifier~$f^{(m)}$, and $c(\mathbf{x}, \mathbf{x}')$ is a predefined manipulation cost (e.g., Mahalanobis distance). The trade-off parameter~$\lambda > 0$ controls the agent's tolerance between acceptance reward and cost.

\subsubsection{Belief Update.}
After observing the decision feedback $y_t$ for submitted input $\mathbf{x}'_t$, the agent updates its belief over the mechanism set $\mathcal{M}$. Specifically, we adopt a soft Bayesian update with exponentiated likelihoods:
\begin{equation}
    \tilde{b}_{t+1}^{(m)} = \frac{b_t^{(m)} \cdot \left( \Pr_{f^{(m)}}(y_t \mid \mathbf{x}'_t) \right)^{\eta}}{\sum_{m' \in \mathcal{M}} b_t^{(m')} \cdot \left( \Pr_{f^{(m')}}(y_t \mid \mathbf{x}'_t) \right)^{\eta}},
    \label{eq:belief_update}
\end{equation}
where $\eta > 0$ is a belief update rate controlling how strongly the agent reacts to new observations. A larger $\eta$ results in faster concentration of belief, while a smaller $\eta$ leads to more conservative updates.
\begin{remark}
    The belief update step is formulated as a Bayesian update, where the likelihood $\Pr_{f^{(m)}}(y_t|\mathbf{x}'_t)$ represents the probability of observing the feedback $y_t$ under each candidate mechanism $m$.
\end{remark}
The whole process of the belief-guided strategic mechanism is illustrated in Algorithm~\ref{alg:pfa}.

\begin{algorithm}[tb]
\caption{Belief-guided Strategic Mechanism}
\label{alg:pfa}
\begin{algorithmic}[1]
\REQUIRE Fairness mechanism set $\mathcal{M}$; total rounds $T$; belief update rate $\eta$
\STATE Initialize agent's belief $\mathbf{b}_0$ over $\mathcal{M}$ (e.g., uniform distribution)
\FOR{$t = 0$ \textbf{to} $T$}
    \STATE \textit{Strategic Manipulation:}
    \STATE Agent selects input $\mathbf{x}'_t$ by maximizing expected utility under current belief $\mathbf{b}_t$ with Eq.~\eqref{belief-guided-m}
    \STATE Agent submits $\mathbf{x}'_t$ and receives decision feedback
    \STATE \textit{Belief Update:}
    \FOR{each $m \in \mathcal{M}$}
        \STATE Update temporary belief $\tilde{b}_{t+1}^{(m)}$ using Eq.~\eqref{eq:belief_update}
    \ENDFOR
\ENDFOR
\RETURN Final belief distribution $\mathbf{b}_{t=T}$
\end{algorithmic}
\end{algorithm}

\begin{figure*}[t]
    \centering
    \begin{subfigure}[b]{0.239\textwidth}
    \centering
    \includegraphics[width=\linewidth]{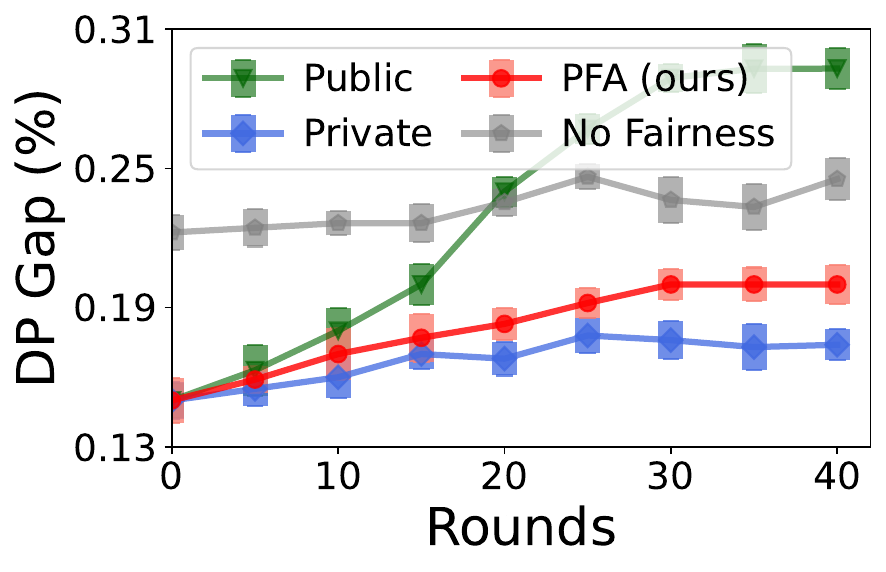}
    \caption{Adult Dataset}
    \label{fig3a}
\end{subfigure}
    \begin{subfigure}[b]{0.239\textwidth}
    \centering
    \includegraphics[width=\linewidth]{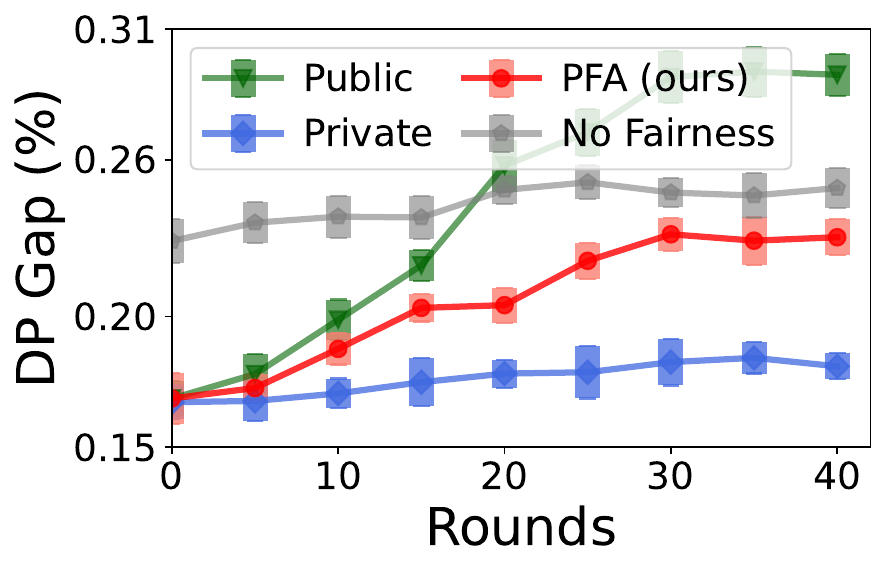}
    \caption{Credit Dataset}
    \label{fig3b}
\end{subfigure}
    \begin{subfigure}[b]{0.239\textwidth}
    \centering
    \includegraphics[width=\linewidth]{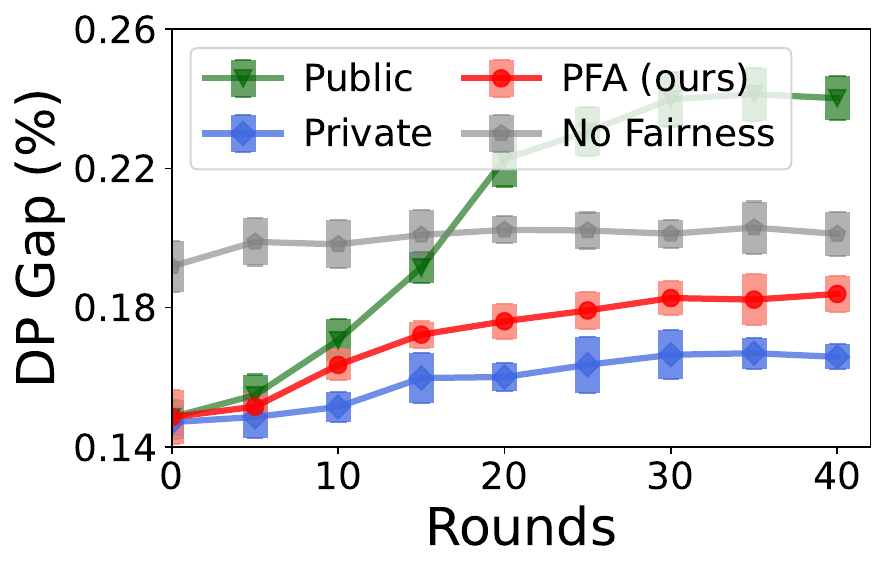}
    \caption{Diabetes Dataset}
    \label{fig3c}
\end{subfigure}
    \begin{subfigure}[b]{0.239\textwidth}
    \centering
    \includegraphics[width=\linewidth]{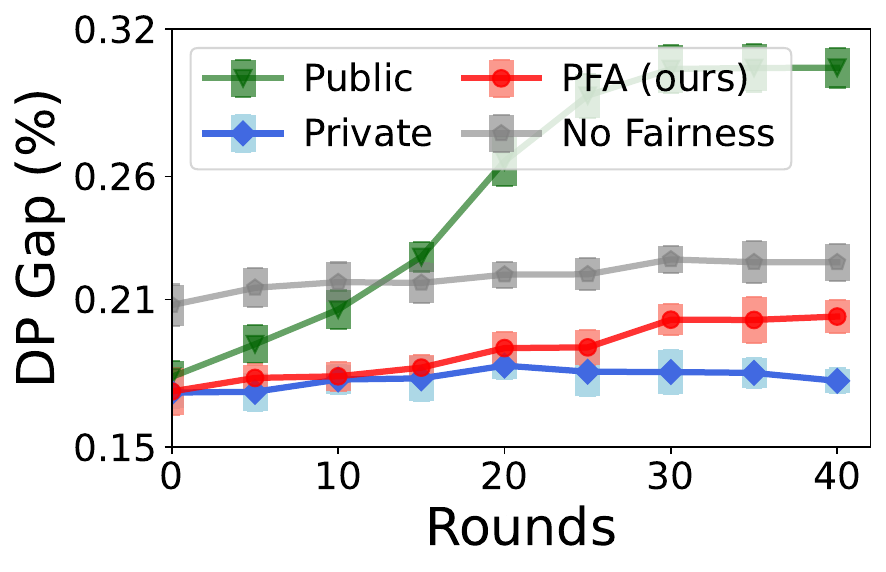}
    \caption{Synthetic Dataset}
    \label{fig3d}
\end{subfigure}
    \caption{Performance of demographic parity gap on different real-world and synthetic datasets.}
    \label{fig3}
\end{figure*}

\begin{figure*}[t]
    \centering
    \begin{subfigure}[b]{0.239\textwidth}
    \centering
    \includegraphics[width=\linewidth]{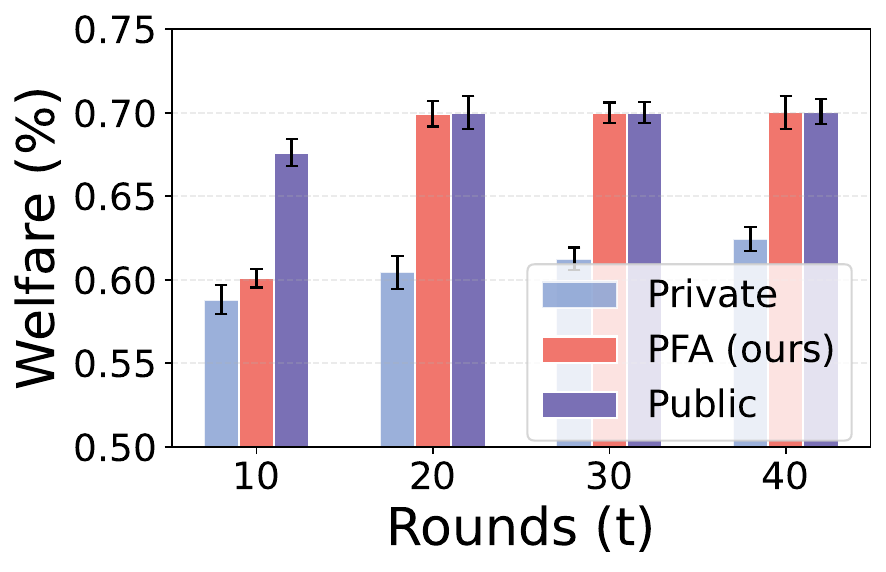}
    \caption{Adult Dataset}
    \label{fig5a}
\end{subfigure}
    \begin{subfigure}[b]{0.239\textwidth}
    \centering
    \includegraphics[width=\linewidth]{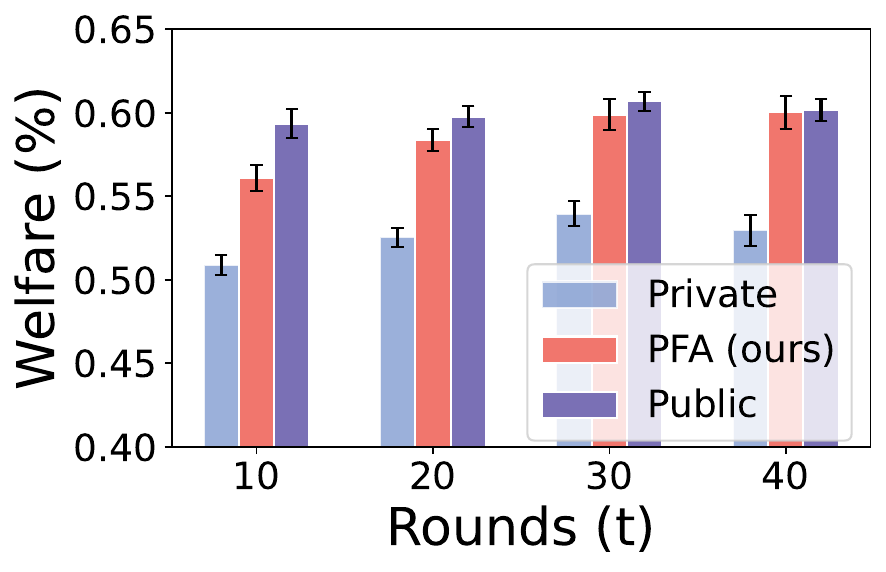}
    \caption{Credit Dataset}
    \label{fig5b}
\end{subfigure}
    \begin{subfigure}[b]{0.239\textwidth}
    \centering
    \includegraphics[width=\linewidth]{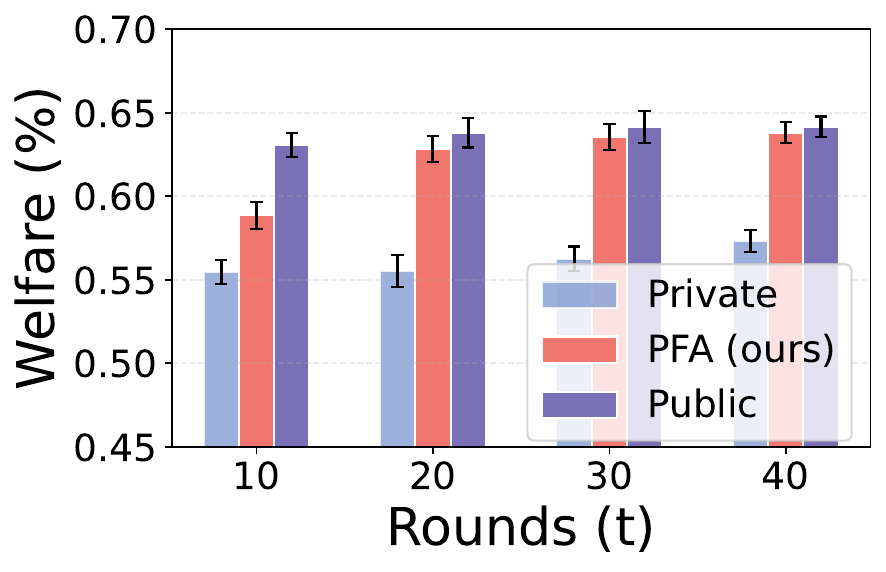}
    \caption{Diabetes Dataset}
    \label{fig5c}
\end{subfigure}
    \begin{subfigure}[b]{0.239\textwidth}
    \centering
    \includegraphics[width=\linewidth]{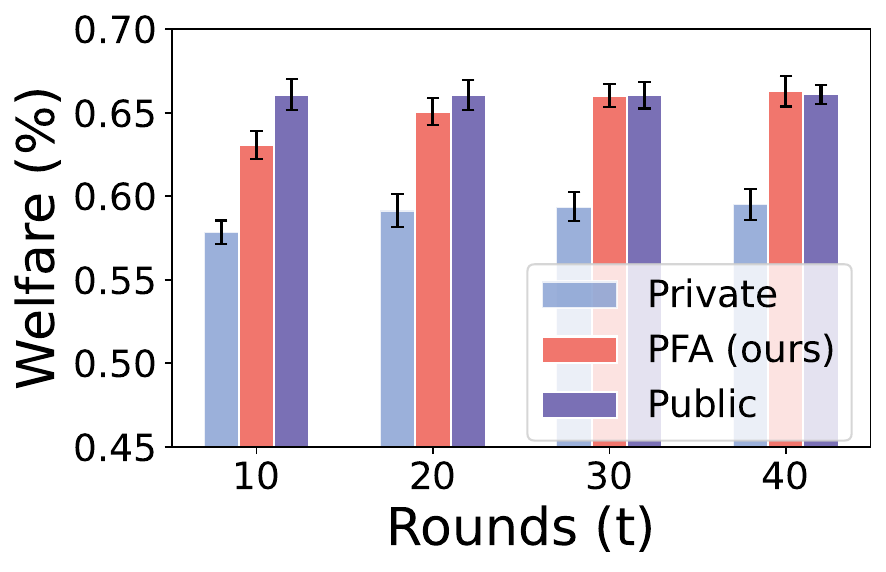}
    \caption{Synthetic Dataset}
    \label{fig5d}
\end{subfigure}
    \caption{Performance of social group welfare on different real-world and synthetic datasets.}
    \label{fig5}
\end{figure*}

\subsection{Theoretical Analysis}

As the agent interacts with the decision system and updates beliefs, provided that the candidate mechanisms in $\mathcal{M}$ are statistically distinguishable,
the agent's belief distribution $\mathbf{b}_t$ is guaranteed to converge to the true mechanism $m^*$. Therefore, we have the following theorem with the proof and convergence rate analysis provided in Appendix C.

\begin{theorem}[Belief Convergence]
Suppose the true mechanism is $m^* \in \mathcal{M}$, and each alternative $m \neq m^*$ is statistically distinguishable based on observed feedback. Then, for any $\epsilon > 0$, there exists $T > 0$ such that:
\begin{equation}
    b_t^{(m^*)} > 1 - \epsilon, \quad \text{with} \ t > T.
\end{equation}
\label{th1}
\end{theorem}

Consequently, the Partial Fairness Awareness (PFA) framework simultaneously addresses two key challenges in strategic classification: it prevents the immediate exploitation of group-specific fairness constraints (thus reducing fairness reversal), and it enables agents to gradually adapt via feedback, increasing the acceptance of truly qualified individuals and improving long-term social welfare. These advantages are formalized below.

\begin{theorem}[Fairness Gap Reduction under PFA\footnote{The specific proofs of Theorem~\ref{th2} and Theorem~\ref{th3} are included in Appendices D and E.}]
Given the fairness gap metric $\Delta_{\text{fair}}$~(e.g., the difference in true positive rates or acceptance rates between protected and advantaged groups). Let $\mathbb{E}[\Delta_{\text{fair}}^{\text{PFA}}]$ and $\mathbb{E}[\Delta_{\text{fair}}^{\text{Public}}]$ be the expected fairness gaps under the Partial Fairness Awareness (PFA) and fully public fairness mechanisms, respectively. Then,
\begin{equation}
    \mathbb{E}[\Delta_{\text{fair}}^{\text{PFA}}] < \mathbb{E}[\Delta_{\text{fair}}^{\text{Public}}].
\end{equation}
\label{th2}
\end{theorem}
\begin{theorem}[Improvement of Social Welfare under PFA]
Let $W_{\text{PFA}}$ and $W_{\text{Private}}$ denote the expected social welfare achieved under the PFA and private fairness settings, respectively, defined as the expected sum of qualified agents accepted minus the cost of manipulation. Then,
\begin{equation}
    \mathbb{E}[W_{\text{PFA}}] > \mathbb{E}[W_{\text{Private}}],
\end{equation}
where the expectation is over agents' sequential learning and adaptation dynamics.
\label{th3}
\end{theorem}

\begin{table}[h]
\centering
\small
\renewcommand{\arraystretch}{1.01}
\begin{tabular}{lcccccc}
\toprule
 \textit{\textbf{DP Gap}(\%)} & $t=5$ & $10$ & $15$ & $20$ & $30$ & $40$ \\
\midrule
 \textit{$\eta=$ 0.01 }& 0.170 & 0.172 & 0.175 & 0.182  & 0.186 & 0.188 \\
 \textit{$\eta=$ 0.05} & 0.173 & 0.178 & 0.182 & 0.189  & 0.192 & 0.192 \\
 \textit{$\eta=$ 0.1 } & 0.181 & 0.183 & 0.186 & 0.190  & 0.193 & 0.193 \\
 \textit{$\eta=$ 0.5}  & 0.182 & 0.185 & 0.189 & 0.193  & 0.195 & 0.196 \\
\bottomrule
\end{tabular}
\caption{Performance of demographic parity gap for different learning rates $\eta$ with rounds $T$.}
\label{tab1}
\end{table}

\begin{table}[h]
\centering
\small
\renewcommand{\arraystretch}{1.01}
\begin{tabular}{lcccccc}
\toprule
\textit{\textbf{Welfare}(\%)} & $t=5$ & $10$ & $15$ & $20$  & $30$ & $40$ \\
\midrule
\textit{$\eta=$ 0.01 } & 0.545 & 0.552 & 0.562 & 0.568  & 0.573 & 0.573 \\
\textit{$\eta=$ 0.05} & 0.550 & 0.561 & 0.568 & 0.573 & 0.580 & 0.581\\
\textit{$\eta=$ 0.1}  & 0.553 & 0.566 & 0.572 & 0.581 & 0.588 & 0.590 \\
\textit{$\eta=$ 0.5} & 0.561 & 0.578 & 0.585 & 0.589 & 0.596 & 0.597 \\
\bottomrule
\end{tabular}
\caption{Performance of the group welfare (\%) for different learning rates $\eta$ with rounds $T$.}
\label{tab2}
\end{table}

\section{Experiment}

\subsection{Experimental Setup}

\noindent \textbf{Datasets.}
We evaluate our framework on four datasets, i.e., three real-world datasets and one synthetic benchmark:
\begin{itemize}
    \item \textit{Credit}~\cite{10.1016/j.eswa.2007.12.020}: Credit card default prediction based on financial records, using gender as the sensitive attribute.
    \item \textit{Adult}~\cite{adult_2}: Income classification from census features; gender is used as the protected attribute.
    \item \textit{Diabetes}~\cite{Teboul2015Diabetes}: A medical dataset containing clinical and demographic attributes used to assess the risk of diabetes, using gender as the sensitive attribute.
    \item \textit{Synthetic}~\cite{lopez2016paysim}: Simulated mobile transaction data for fraud detection, with group membership as the sensitive attribute.
\end{itemize}

\noindent \textbf{Metrics.}
We report fairness metrics and the group welfare metric. In fairness metrics, we mainly use the DP~(demographic parity) Gap, but also employ EO~(equalized odds) gap and PP~(predictive parity) gap for more comprehensive evaluation.
\begin{itemize}
    \item \textbf{DP Gap} measures the disparity in acceptance rates across groups:
    \begin{equation}
        \mathrm{DP\ Gap} = \left| \mathbb{P}(\hat{y} = 1 \mid g = A) - \mathbb{P}(\hat{y} = 1 \mid g = B) \right|,
    \end{equation}
    where lower values indicate better statistical parity.
    \item \textbf{Group welfare}, defined as in Eq.~\eqref{welfare}, measuring the acceptance of truly qualified individuals across groups.
\end{itemize}

\begin{figure}[t]
	\centering
    \begin{subfigure}{0.49\linewidth}
    \centering
    \includegraphics[width=0.49\linewidth]{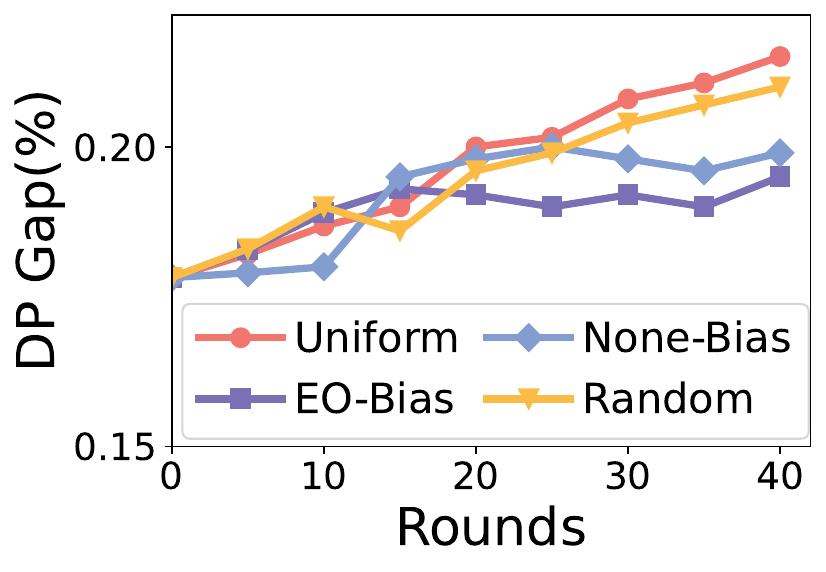}
    \caption{DP Gap}
    \label{fig8a}
\end{subfigure}
    \begin{subfigure}{0.49\linewidth}
    \centering
    \includegraphics[width=0.49\linewidth]{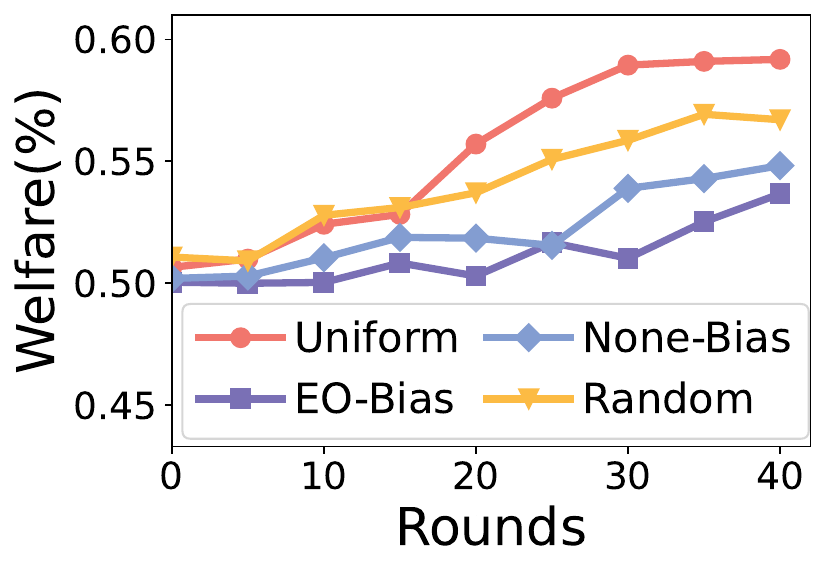}
    \caption{Group Welfare}
    \label{fig8b}
\end{subfigure}
    \caption{Ablation experimental results of the belief initialization distribution.}
    \label{fig8}
\end{figure}

\begin{figure}[t]
	\centering
	\begin{subfigure}{0.479\linewidth}
    \centering
    \includegraphics[width=0.479\linewidth]{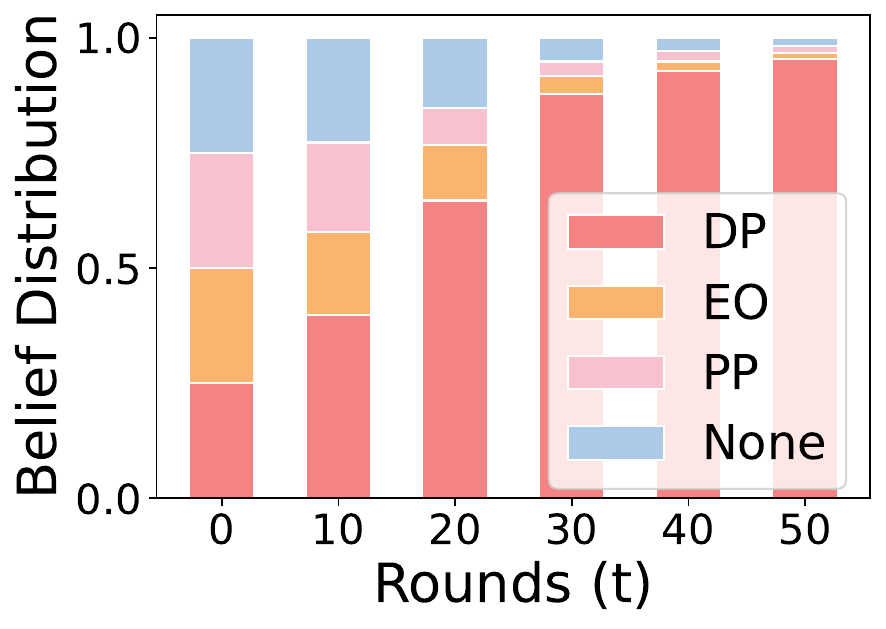}
    \caption{Uniform Initialization}
    \label{fig9a}
\end{subfigure}
    \begin{subfigure}{0.479\linewidth}
    \centering
    \includegraphics[width=0.479\linewidth]{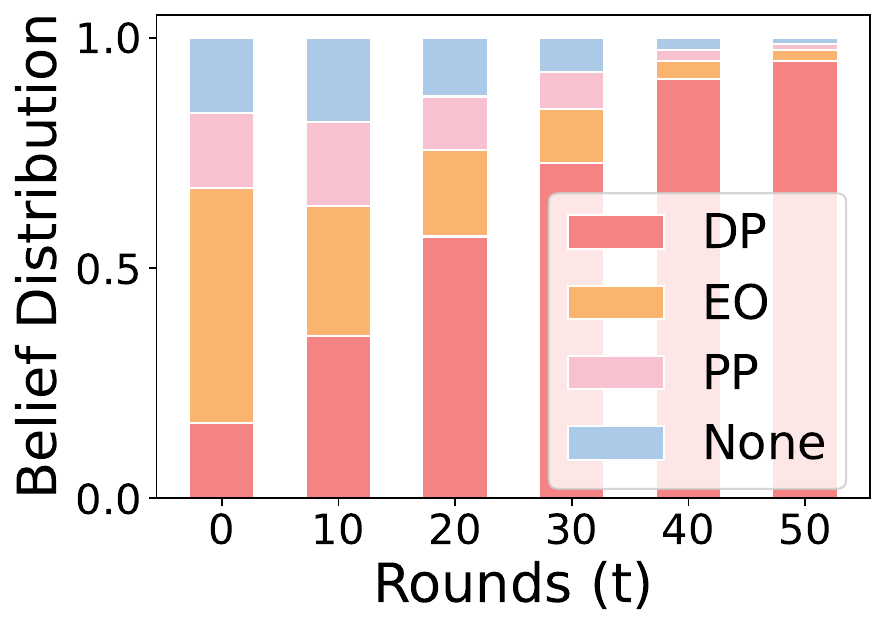}
    \caption{EO-biased Initialization}
    \label{fig9b}
\end{subfigure}
    \caption{Ablation experimental results of the belief initialization distribution.}
    \label{fig9}
\end{figure}

\noindent \textbf{Fairness Settings.}
We compare three fairness strategies: (i)\textbf{public fairness}, where group-specific constraints are fully disclosed to agents. (ii)\textbf{private fairness}, where all constraints are hidden and agents assume a single threshold. (iii) \textbf{Partial Fairness Awareness}, where agents gradually learn the underlying mechanism from decision feedback over multiple rounds.

\noindent \textbf{Baseline.}
All mechanisms use the same linear classifier and Mahalanobis distance for manipulation cost~\cite{gavish2021optimalrecoveryprecisionmatrix}, ensuring only the transparency level differs. To evaluate alignment between classifier decisions and agent qualification, we use the \emph{strategic improvement} framework~\cite{miller2020strategic,chen2023learning}. The experimental results are presented in Figures~\ref{fig3} and~\ref{fig5}.

\subsection{Implementation Details}

All experiments are implemented in Python 3.10 using the \textit{scikit-learn} library for linear classifiers. The manipulation cost is computed as the Mahalanobis distance~\cite{gavish2021optimalrecoveryprecisionmatrix}, consistent across all settings. For PFA, belief updates use a multiplicative weights algorithm with learning rate $\eta = 0.002$. Each agent interacts with the classifier for up to $T = 40$ rounds. All experiments are conducted on a single NVIDIA TITAN V (12GB) GPU. More details and experimental results are provided in Appendix F.

\subsection{Ablation Study}
To investigate the effect of belief-guided learning in PFA, we conduct two ablation studies as follows. Unless otherwise stated, all other parameters are fixed as above.

\noindent \textbf{Belief Update Rate.}
We vary the belief update rate \(\eta\) in $\{0.01, 0.05, 0.1, 0.5\}$ to evaluate its effect on convergence and adaptation stability. Beliefs are initialized uniformly over the candidate mechanism set \(\mathcal{M}\), and agents interact for $T=40$ rounds. Results are in Tables~\ref{tab1} and~\ref{tab2}.

\noindent \textbf{Initial Belief Distribution.}
To assess robustness to prior knowledge or bias, we experiment with non-uniform initial beliefs, including strong priors (e.g., EO-biased $b_{t=0}^{(\text{EO})}=0.51$) and random weightings across $\mathcal{M}$. All other settings remain constant. Results are summarized in Figure~\ref{fig8}.

\subsection{Result Analysis}

\noindent \textbf{Main Results. } Figure~\ref{fig3} summarizes the group fairness outcomes under different mechanisms, as measured by the Demographic Parity (DP) gap, across four representative datasets. Consistently, the \textbf{PFA (ours)} method achieves a markedly lower group-fairness gap compared to the public fairness. Notably, this improvement is robust across both real-world datasets (Adult, Credit, Diabetes) and the synthetic benchmark. As the number of interaction rounds increases, the DP gap for PFA remains stable and does not exhibit the sharp increases seen under Public or No Fairness settings, especially after $T>20$, indicating strong resistance to fairness reversal. These findings provide empirical validation for Theorem~\ref{th2}.

Figure~\ref{fig5} further illustrates the corresponding dynamics of social group welfare under the same settings. Here, the \textbf{PFA (ours)} not only matches but often surpasses the welfare attained by the Public baseline, particularly in multiple rounds. In all datasets, social welfare improves steadily with additional rounds of interaction under PFA. This trend confirms that PFA enables agents to adapt their behavior in alignment with both fairness and qualification, effectively mitigating the typical welfare loss associated with private fairness constraints. These observations support Theorem~\ref{th3}.

As illustrated in Figure~\ref{fig9}, we examine the evolution of agent beliefs under different initializations. In all cases, the belief distribution quickly converges toward the true fairness mechanism (DP) adopted by the jury, regardless of the initial setting. This empirical result aligns with Theorem~\ref{th1}, confirming that agents can reliably recover the underlying mechanism after a moderate number of interaction rounds, even when starting from a uniform or biased prior.

\noindent \textbf{Ablation Results. }
The ablation studies evaluate the sensitivity of the PFA framework to two key factors: the belief update rate $\eta$ and the initial belief distribution. As shown in Tables~\ref{tab1} and \ref{tab2}, intermediate learning rates ($\eta = 0.05$ or $0.1$) achieve the best balance, yielding both low group fairness gaps and high social welfare. In contrast, very small learning rates slow down convergence, while overly large rates can introduce instability and degrade early performance.

Figure~\ref{fig8} further examines the effect of varying initial belief distributions. Despite some transient fluctuations in fairness metrics and welfare during early rounds, the belief consistently converges to the correct fairness mechanism and achieves comparable welfare across all settings. These findings demonstrate the robustness of the belief-guided mechanism to prior uncertainty in the initial belief distribution.

Overall, these results demonstrate that the PFA strikes a favorable balance between fairness and social welfare, maintaining equity across groups while supporting qualified individuals.

\section{Conclusion}

This work formulates the Partial Fairness Awareness (PFA) problem, addressing the fundamental trade-off between public and private fairness constraints in strategic classification. To solve this problem, we develop a belief-guided strategic mechanism that enables agents to update their belief distribution of the fairness mechanism set based on feedback from the decision system. Theoretical analysis and experimental results on real-world and synthetic datasets demonstrate that PFA effectively mitigates fairness reversal and enhances social welfare in strategic machine learning.
Future directions include extending PFA to multi-class settings, modeling heterogeneous belief-update dynamics, and adapting to dynamic environments.

\section*{Ethics Statement}
This work does not raise any ethical concerns.
All experiments are conducted on publicly available datasets, and no human subjects or sensitive attributes are involved.

\section*{Acknowledgments}

This work is supported in part by the National Natural Science Foundation of China under Grants No.62372459 and 62525213, in part by the Shanghai Sailing Program under Grant No. 24YF2711600, and in part by the NUDT Youth Independent Innovation Science Fund under Grant No. ZK25-20.

\bibliographystyle{plainnat}
\bibliography{aaai2026}

@inproceedings{sanchez2020does,
  title={What does it mean to'solve'the problem of discrimination in hiring? Social, technical and legal perspectives from the UK on automated hiring systems},
  author={S{\'a}nchez-Monedero, Javier and Dencik, Lina and Edwards, Lilian},
  booktitle={Proceedings of the 2020 conference on fairness, accountability, and transparency},
  pages={458--468},
  year={2020}
}

@article{shimao2025strategic,
  title={Strategic Best-Response Fairness Framework for Fair Machine Learning},
  author={Shimao, Hajime and Khern-Am-Nuai, Warut and Kannan, Karthik and Cohen, Maxime C},
  journal={Information Systems Research},
  year={2025},
  publisher={INFORMS}
}

@misc{gavish2021optimalrecoveryprecisionmatrix,
      title={Optimal Recovery of Precision Matrix for Mahalanobis Distance from High Dimensional Noisy Observations in Manifold Learning}, 
      author={Matan Gavish and Ronen Talmon and Pei-Chun Su and Hau-Tieng Wu},
      year={2021},
      eprint={1904.09204},
      archivePrefix={arXiv},
      primaryClass={math.ST},
      url={https://arxiv.org/abs/1904.09204}, 
}

@article{bent2019algorithmic,
  title={Is algorithmic affirmative action legal},
  author={Bent, Jason R},
  journal={Geo. LJ},
  volume={108},
  pages={803},
  year={2019},
  publisher={HeinOnline}
}

@article{Strathern_1997, title={‘Improving ratings’: audit in the British University system}, volume={5}, DOI={10.1002/(SICI)1234-981X(199707)5:3&#60;305::AID-EURO184&#62;3.0.CO;2-4}, number={3}, journal={European Review}, author={Strathern, Marilyn}, year={1997}, pages={305–321}}

@inproceedings{mutlu2022contrastive,
  title={Contrastive counterfactual fairness in algorithmic decision-making},
  author={Mutlu, Ece {\c{C}}i{\u{g}}dem and Yousefi, Niloofar and Ozmen Garibay, Ozlem},
  booktitle={Proceedings of the 2022 AAAI/ACM Conference on AI, Ethics, and Society},
  pages={499--507},
  year={2022}
}

@misc{Teboul2015Diabetes,
  author = {Alex Teboul},
  title = {Diabetes Health Indicators Dataset},
  year = {2015},
  url = {https://www.kaggle.com/datasets/alexteboul/diabetes-health-indicators-dataset}
}

@inproceedings{fairnessreverse,
author = {Estornell, Andrew and Das, Sanmay and Liu, Yang and Vorobeychik, Yevgeniy},
title = {Group-Fair Classification with Strategic Agents},
year = {2023},
isbn = {9798400701924},
publisher = {Association for Computing Machinery},
address = {New York, NY, USA},
url = {https://doi.org/10.1145/3593013.3594006},
doi = {10.1145/3593013.3594006},
booktitle = {Proceedings of the 2023 ACM Conference on Fairness, Accountability, and Transparency},
pages = {389–399},
numpages = {11},
location = {Chicago, IL, USA},
series = {FAccT '23}
}

@article{lv2026tabular,
  title={When Tabular Foundation Models Meet Strategic Tabular Data: A Prior Alignment Approach},
  author={Lv, Xinpeng and Mao, Yunxin and Xu, Renzhe and Zheng, Chunyuan and Chen, Yikai and Li, Haoxuan and Yang, Jinxuan and Kuang, Kun and Chen, Yuanlong and Geng, Mingyang and others},
  journal={arXiv preprint arXiv:2605.19662},
  year={2026}
}

@article{lv2026breaking,
  title={Breaking the Gradient Barrier: Unveiling Large Language Models for Strategic Classification},
  author={Lv, Xinpeng and Mao, Yunxin and Li, Haoxuan and Liang, Ke and Yang, Jinxuan and Huang, Wanrong and Chi, Haoang and Chen, Huan and Lan, Long and Yang, Wenjing and others},
  journal={Advances in Neural Information Processing Systems},
  volume={38},
  pages={49145--49180},
  year={2026}
}

@article{lv2026beyond,
  title={Beyond Rational Illusion: Behaviorally Realistic Strategic Classification},
  author={Lv, Xinpeng and Mao, Yunxin and Xu, Renzhe and Zheng, Chunyuan and Chen, Yikai and Li, Haoxuan and Shi, Yang and Yang, Jinxuan and Lin, Zhouchen and Chen, Yuanlong and others},
  journal={arXiv preprint arXiv:2605.19674},
  year={2026}
}

@article{kuvcak2018machine,
  title={MACHINE LEARNING IN EDUCATION-A SURVEY OF CURRENT RESEARCH TRENDS.},
  author={Kuvcak, Danijel and Jurivcic, Vedran and DJambic, Goran},
  journal={Annals of DAAAM \& Proceedings},
  volume={29},
  year={2018}
}

@inproceedings{hu2019disparate,
  title={The disparate effects of strategic manipulation},
  author={Hu, Lily and Immorlica, Nicole and Vaughan, Jennifer Wortman},
  booktitle={Proceedings of the Conference on Fairness, Accountability, and Transparency},
  pages={259--268},
  year={2019}
}

@misc{dong2017strategicclassificationrevealedpreferences,
      title={Strategic Classification from Revealed Preferences}, 
      author={Jinshuo Dong and Aaron Roth and Zachary Schutzman and Bo Waggoner and Zhiwei Steven Wu},
      year={2017},
      eprint={1710.07887},
      archivePrefix={arXiv},
      primaryClass={cs.LG},
      url={https://arxiv.org/abs/1710.07887}, 
}

@article{chen2020learning,
  title={Learning strategy-aware linear classifiers},
  author={Chen, Yiling and Liu, Yang and Podimata, Chara},
  journal={Advances in Neural Information Processing Systems},
  volume={33},
  pages={15265--15276},
  year={2020}
}

@article{jagtiani2019roles,
  title={The roles of alternative data and machine learning in fintech lending: evidence from the LendingClub consumer platform},
  author={Jagtiani, Julapa and Lemieux, Catharine},
  journal={Financial Management},
  volume={48},
  number={4},
  pages={1009--1029},
  year={2019},
  publisher={Wiley Online Library}
}

@inproceedings{NEURIPS2021_f1404c26,
 author = {Harris, Keegan and Heidari, Hoda and Wu, Steven Z.},
 booktitle = {Advances in Neural Information Processing Systems},
 editor = {M. Ranzato and A. Beygelzimer and Y. Dauphin and P.S. Liang and J. Wortman Vaughan},
 pages = {28728--28741},
 publisher = {Curran Associates, Inc.},
 title = {Stateful Strategic Regression},
 url = {https://proceedings.neurips.cc/paper_files/paper/2021/file/f1404c2624fa7f2507ba04fd9dfc5fb1-Paper.pdf},
 volume = {34},
 year = {2021}

}

@article{hardt2023performative,
  title={Performative prediction: Past and future},
  author={Hardt, Moritz and Mendler-D{\"u}nner, Celestine},
  journal={arXiv preprint arXiv:2310.16608},
  year={2023}
}

@inproceedings{perdomo2020performative,
  title={Performative prediction},
  author={Perdomo, Juan and Zrnic, Tijana and Mendler-D{\"u}nner, Celestine and Hardt, Moritz},
  booktitle={International Conference on Machine Learning},
  pages={7599--7609},
  year={2020},
  organization={PMLR}
}

@article{hardt2022performative,
  title={Performative power},
  author={Hardt, Moritz and Jagadeesan, Meena and Mendler-D{\"u}nner, Celestine},
  journal={Advances in Neural Information Processing Systems},
  volume={35},
  pages={22969--22981},
  year={2022}
}

@inproceedings{hardt2016strategic,
  title={Strategic classification},
  author={Hardt, Moritz and Megiddo, Nimrod and Papadimitriou, Christos and Wootters, Mary},
  booktitle={Proceedings of the 2016 ACM conference on innovations in theoretical computer science},
  pages={111--122},
  year={2016}
}

@inproceedings{milli2019social,
  title={The social cost of strategic classification},
  author={Milli, Smitha and Miller, John and Dragan, Anca D and Hardt, Moritz},
  booktitle={Proceedings of the Conference on Fairness, Accountability, and Transparency},
  pages={230--239},
  year={2019}
}

@inproceedings{miller2020strategic,
  title={Strategic classification is causal modeling in disguise},
  author={Miller, John and Milli, Smitha and Hardt, Moritz},
  booktitle={International Conference on Machine Learning},
  pages={6917--6926},
  year={2020},
  organization={PMLR}
}

@article{mendler2022anticipating,
  title={Anticipating performativity by predicting from predictions},
  author={Mendler-D{\"u}nner, Celestine and Ding, Frances and Wang, Yixin},
  journal={Advances in neural information processing systems},
  volume={35},
  pages={31171--31185},
  year={2022}
}

@article{haghtalab2020maximizing,
  title={Maximizing welfare with incentive-aware evaluation mechanisms},
  author={Haghtalab, Nika and Immorlica, Nicole and Lucier, Brendan and Wang, Jack Z},
  journal={arXiv preprint arXiv:2011.01956},
  year={2020}
}

@article{zrnic2021leads,
  title={Who leads and who follows in strategic classification?},
  author={Zrnic, Tijana and Mazumdar, Eric and Sastry, Shankar and Jordan, Michael},
  journal={Advances in Neural Information Processing Systems},
  volume={34},
  pages={15257--15269},
  year={2021}
}

@article{
chen2023learning,
title={Learning to Incentivize Improvements from Strategic Agents },
author={Yatong Chen and Jialu Wang and Yang Liu},
journal={Transactions on Machine Learning Research},
issn={2835-8856},
year={2023},
url={https://openreview.net/forum?id=W98AEKQ38Y},
note={}
}

@incollection{roemer2015equality,
  title={Equality of opportunity},
  author={Roemer, John E and Trannoy, Alain},
  booktitle={Handbook of income distribution},
  volume={2},
  pages={217--300},
  year={2015},
  publisher={Elsevier}
}

@article{dieterich2016compas,
  title={COMPAS risk scales: Demonstrating accuracy equity and predictive parity},
  author={Dieterich, William and Mendoza, Christina and Brennan, Tim},
  journal={Northpointe Inc},
  volume={7},
  number={4},
  pages={1--36},
  year={2016}
}

@inproceedings{zemel2013learning,
  title={Learning Fair Representations},
  author={Zemel, Richard and Wu, Yu and Swersky, Kevin and Pitassi, Toniann and Dwork, Cynthia},
  booktitle={International Conference on Machine Learning},
  pages={325--333},
  year={2013}
}

@article{shao2024strategic,
  title={Strategic classification under unknown personalized manipulation},
  author={Shao, Han and Blum, Avrim and Montasser, Omar},
  journal={Advances in Neural Information Processing Systems},
  volume={36},
  year={2024}
}

@article{li2017review,
  title={A REVIEW OF DYNAMIC STACKELBERG GAME MODELS.},
  author={Li, Tao and Sethi, Suresh P},
  journal={Discrete \& Continuous Dynamical Systems-Series B},
  volume={22},
  number={1},
  year={2017}
}

@inproceedings{horowitz2023causal,
  title={Causal strategic classification: A tale of two shifts},
  author={Horowitz, Guy and Rosenfeld, Nir},
  booktitle={International Conference on Machine Learning},
  pages={13233--13253},
  year={2023},
  organization={PMLR}
}

@inproceedings{shavit2020causal,
  title={Causal strategic linear regression},
  author={Shavit, Yonadav and Edelman, Benjamin and Axelrod, Brian},
  booktitle={International Conference on Machine Learning},
  pages={8676--8686},
  year={2020},
  organization={PMLR}
}

@inproceedings{keswani2023addressing,
  title={Addressing strategic manipulation disparities in fair classification},
  author={Keswani, Vijay and Celis, L Elisa},
  booktitle={Proceedings of the 3rd ACM Conference on Equity and Access in Algorithms, Mechanisms, and Optimization},
  pages={1--11},
  year={2023}
}

@InProceedings{10.1007/978-3-032-05962-8_25,
author="Yang, Wenjing
and Lv, Xinpeng
and Mao, Yunxin
and Xu, Liyang
and Jin, Ruochun
and Chen, Huan
and Ren, Jing
and Yang, Jinxuan
and Chen, Yuanlong
and Wang, Haotian",
title="Advanced Strategic Improvement with Decision Interactions",
booktitle="Machine Learning and Knowledge Discovery in Databases. Research Track",
year="2026",
publisher="Springer Nature Switzerland",
address="Cham",
pages="426--444",
isbn="978-3-032-05962-8"
}

@ARTICLE{11097075,
  author={Yang, Songyuan and Yu, Weijiang and Yang, Wenjing and Liu, Xinwang and Tan, Huibin and Lan, Long and Xiao, Nong},
  journal={IEEE Transactions on Pattern Analysis and Machine Intelligence}, 
  title={WildVideo: Benchmarking LMMs for Understanding Video-Language Interaction}, 
  year={2025},
  volume={47},
  number={10},
  pages={9330-9344},
  doi={10.1109/TPAMI.2025.3592831}}

@inproceedings{zhengchunyuan,
author = {Zheng, Chunyuan and Pan, Hang and Zhang, Yang and Li, Haoxuan},
title = {Adaptive Structure Learning with Partial Parameter Sharing for Post-Click Conversion Rate Prediction},
year = {2025},
isbn = {9798400715921},
publisher = {Association for Computing Machinery},
address = {New York, NY, USA},
url = {https://doi.org/10.1145/3726302.3729887},
doi = {10.1145/3726302.3729887},
booktitle = {Proceedings of the 48th International ACM SIGIR Conference on Research and Development in Information Retrieval},
pages = {233–243},
numpages = {11},
location = {Padua, Italy},
series = {SIGIR '25}
}

@inproceedings{wang2025simprof,
  title={SimProF: A Simple Probabilistic Framework for Unsupervised Domain Adaptation},
  author={Wang, Meng-zhu},
  booktitle={Proceedings of the AAAI Conference on Artificial Intelligence},
  volume={39},
  number={20},
  pages={21153--21161},
  year={2025}
}

@inproceedings{yang2024your,
  title={Your neighbor matters: Towards fair decisions under networked interference},
  author={Yang, Wenjing and Wang, Haotian and Li, Haoxuan and Zou, Hao and Jin, Ruochun and Kuang, Kun and Cui, Peng},
  booktitle={Proceedings of the 30th ACM SIGKDD Conference on Knowledge Discovery and Data Mining},
  pages={3829--3840},
  year={2024}
}

@inproceedings{wang2022estimating,
  title={Estimating individualized causal effect with confounded instruments},
  author={Wang, Haotian and Yang, Wenjing and Yang, Longqi and Wu, Anpeng and Xu, Liyang and Ren, Jing and Wu, Fei and Kuang, Kun},
  booktitle={Proceedings of the 28th ACM SIGKDD Conference on Knowledge Discovery and Data Mining},
  pages={1857--1867},
  year={2022}
}

@inproceedings{wang2023treatment,
  title={Treatment effect estimation with adjustment feature selection},
  author={Wang, Haotian and Kuang, Kun and Chi, Haoang and Yang, Longqi and Geng, Mingyang and Huang, Wanrong and Yang, Wenjing},
  booktitle={Proceedings of the 29th ACM SIGKDD Conference on Knowledge Discovery and Data Mining},
  pages={2290--2301},
  year={2023}
}

@article{diana2024minimax,
  title={Minimax Group Fairness in Strategic Classification},
  author={Diana, Emily and Sharifi-Malvajerdi, Saeed and Vakilian, Ali},
  journal={arXiv preprint arXiv:2410.02513},
  year={2024}
}

@article{alhanouti2025anticipating,
  title={Anticipating Gaming to Incentivize Improvement: Guiding Agents in (Fair) Strategic Classification},
  author={Alhanouti, Sura and Naghizadeh, Parinaz},
  journal={arXiv preprint arXiv:2505.05594},
  year={2025}
}

@inproceedings{zhang2022fairness,
  title={Fairness interventions as (dis) incentives for strategic manipulation},
  author={Zhang, Xueru and Khalili, Mohammad Mahdi and Jin, Kun and Naghizadeh, Parinaz and Liu, Mingyan},
  booktitle={International Conference on Machine Learning},
  pages={26239--26264},
  year={2022},
  organization={PMLR}
}

@article{tsirtsis2024optimal,
  title={Optimal decision making under strategic behavior},
  author={Tsirtsis, Stratis and Tabibian, Behzad and Khajehnejad, Moein and Singla, Adish and Sch{\"o}lkopf, Bernhard and Gomez-Rodriguez, Manuel},
  journal={Management Science},
  year={2024},
  publisher={INFORMS}
}

@inproceedings{mofakhami2023performative,
  title={Performative prediction with neural networks},
  author={Mofakhami, Mehrnaz and Mitliagkas, Ioannis and Gidel, Gauthier},
  booktitle={International Conference on Artificial Intelligence and Statistics},
  pages={11079--11093},
  year={2023},
  organization={PMLR}
}

@inproceedings{vo2024causal,
  title={Causal Strategic Learning with Competitive Selection},
  author={Vo, Kiet QH and Aadil, Muneeb and Chau, Siu Lun and Muandet, Krikamol},
  booktitle={Proceedings of the AAAI Conference on Artificial Intelligence},
  volume={38},
  pages={15411--15419},
  year={2024}
}

@article{efthymiou2025incentivizing,
  title={Incentivizing Desirable Effort Profiles in Strategic Classification: The Role of Causality and Uncertainty},
  author={Efthymiou, Valia and Podimata, Chara and Sen, Diptangshu and Ziani, Juba},
  journal={arXiv preprint arXiv:2502.06749},
  year={2025}
}

@article{chang2024s,
  title={Who’s gaming the system? a causally-motivated approach for detecting strategic adaptation},
  author={Chang, Trenton and Warrenburg, Lindsay and Park, Sae-Hwan and Parikh, Ravi and Makar, Maggie and Wiens, Jenna},
  journal={Advances in Neural Information Processing Systems},
  volume={37},
  pages={42311--42348},
  year={2024}
}

@inproceedings{estornell2023incentivizing,
  title={Incentivizing recourse through auditing in strategic classification},
  author={Estornell, Andrew and Chen, Yatong and Das, Sanmay and Liu, Yang and Vorobeychik, Yevgeniy},
  booktitle={IJCAI},
  year={2023}
}

@inproceedings{xie2024non,
  title={Non-linear welfare-aware strategic learning},
  author={Xie, Tian and Zhang, Xueru},
  booktitle={Proceedings of the AAAI/ACM Conference on AI, Ethics, and Society},
  volume={7},
  pages={1660--1671},
  year={2024}
}

@misc{adult_2,
  author       = {Becker, Barry and Kohavi, Ronny},
  title        = {{Adult}},
  year         = {1996},
  howpublished = {UCI Machine Learning Repository},
  note         = {{DOI}: https://doi.org/10.24432/C5XW20}
}

@InProceedings{pmlr-v139-ghalme21a,
  title = 	 {Strategic Classification in the Dark},
  author =       {Ghalme, Ganesh and Nair, Vineet and Eilat, Itay and Talgam-Cohen, Inbal and Rosenfeld, Nir},
  booktitle = 	 {Proceedings of the 38th International Conference on Machine Learning},
  pages = 	 {3672--3681},
  year = 	 {2021},
  editor = 	 {Meila, Marina and Zhang, Tong},
  volume = 	 {139},
  series = 	 {Proceedings of Machine Learning Research},
  month = 	 {18--24 Jul},
  publisher =    {PMLR},
  url = 	 {https://proceedings.mlr.press/v139/ghalme21a.html}
}

@article{10.1016/j.eswa.2007.12.020,
author = {Yeh, I-Cheng and Lien, Che-hui},
title = {The Comparisons of Data Mining Techniques for the Predictive Accuracy of Probability of Default of Credit Card Clients},
year = {2009},
issue_date = {March, 2009},
publisher = {Pergamon Press, Inc.},
address = {USA},
volume = {36},
number = {2},
issn = {0957-4174},
url = {https://doi.org/10.1016/j.eswa.2007.12.020},
doi = {10.1016/j.eswa.2007.12.020},
journal = {Expert Syst. Appl.},
month = {mar},
pages = {2473–2480},
numpages = {8}
}

@inproceedings{singh2024optimal,
  title={Optimal Stochastic Decision Rule for Strategic Classification},
  author={Singh, Manish Kumar and Kulkarni, Ankur A},
  booktitle={2024 National Conference on Communications (NCC)},
  pages={1--6},
  year={2024},
  organization={IEEE}
}

@inproceedings{lopez2016paysim,
  title={PaySim: A financial mobile money simulator for fraud detection},
  author={Lopez-Rojas, Edgar and Elmir, Ahmad and Axelsson, Stefan},
  booktitle={28th European modeling and simulation symposium, EMSS, Larnaca},
  pages={249--255},
  year={2016},
  organization={Dime University of Genoa}
}

\appendix

\section{A. Proof of Theorem 1}

We formally analyze the social welfare gap under different fairness constraint visibility regimes in strategic classification. To do so, we clarify the types of group-specific fairness constraints, model the agent's strategic response, and quantify the resulting welfare difference.

\subsection{Preliminaries and Notation}
Let $\mathbf{x} \in \mathbb{R}^d$ denote the raw feature vector of an agent, and $\mathbf{x}' = \mathbf{x} + \mathbf{M}\mathbf{a}$ denote the manipulated features, where $\mathbf{M}$ is a transformation matrix and $\mathbf{a}$ the manipulation direction. The cost of manipulation is measured by the Mahalanobis distance:
\begin{equation}
C(\mathbf{a}) = \tfrac{1}{2}\mathbf{a}^\top\mathbf{\Sigma}^{-1}\mathbf{a},
\end{equation}
where $\mathbf{\Sigma}$ is positive definite and encodes effort constraints.

The classifier applies a scoring function $s = \boldsymbol{\omega}^\top\mathbf{V}\mathbf{x}'$, which is compared to a group-dependent threshold $\theta_g$. The agent's true qualification is defined as $y = \boldsymbol{\omega}_*^\top\mathbf{x}' + \eta$, where $\eta \sim \mathcal{N}(0, \sigma^2)$.

We assume group fairness is implemented as threshold adjustments:
\begin{equation}
\theta_a > \theta_o > \theta_b,
\end{equation}
where $\theta_a$ and $\theta_b$ are the thresholds for advantaged and disadvantaged groups, and $\theta_o$ is the baseline (group-agnostic) threshold.

An agent contributes to social welfare if and only if: (i) the agent is truly qualified ($y \geq y_{\text{th}}$), and (ii) is accepted by the classifier ($s \geq \theta_g$).

\subsection{Strategic Response Under Different Regimes}

\paragraph{Scenario 1: Public Fairness.}
Agents know their group $g$ and the corresponding threshold $\theta_g$. The optimal manipulation solves
\begin{equation}
\min_{\mathbf{a}} \ \tfrac{1}{2}\mathbf{a}^\top\mathbf{\Sigma}^{-1}\mathbf{a}
\quad \text{s.t.} \quad
\boldsymbol{\omega}^\top\mathbf{V}(\mathbf{x} + \mathbf{M}\mathbf{a}) = \theta_g.
\end{equation}
The closed-form solution is
\begin{equation}
\mathbf{a}_1^* = \frac{\theta_g - \boldsymbol{\omega}^\top\mathbf{V}\mathbf{x}}{\boldsymbol{\omega}^\top\mathbf{V}\mathbf{M}\mathbf{\Sigma}\mathbf{M}^\top\mathbf{V}\boldsymbol{\omega}} \cdot \mathbf{\Sigma}\mathbf{M}^\top\mathbf{V}\boldsymbol{\omega}.
\end{equation}
The agent's post-manipulation qualification is
\begin{equation}
y^{(1)} = \boldsymbol{\omega}_*^\top\mathbf{x} + \alpha(\theta_g - \boldsymbol{\omega}^\top\mathbf{V}\mathbf{x}) + \eta,
\end{equation}
where $\alpha = \frac{\boldsymbol{\omega}_*^\top\mathbf{M}\mathbf{\Sigma}\mathbf{M}^\top\mathbf{V}\boldsymbol{\omega}}{\boldsymbol{\omega}^\top\mathbf{V}\mathbf{M}\mathbf{\Sigma}\mathbf{M}^\top\mathbf{V}\boldsymbol{\omega}}$.

\paragraph{Scenario 2: Private Fairness.}
Agents do not observe their group-specific threshold and optimize under the assumption of a shared threshold $\theta_o$:
\begin{equation}
\mathbf{a}_2^* = \frac{\theta_o - \boldsymbol{\omega}^\top\mathbf{V}\mathbf{x}}{\boldsymbol{\omega}^\top\mathbf{V}\mathbf{M}\mathbf{\Sigma}\mathbf{M}^\top\mathbf{V}\boldsymbol{\omega}} \cdot \mathbf{\Sigma}\mathbf{M}^\top\mathbf{V}\boldsymbol{\omega},
\end{equation}
with resulting qualification
\begin{equation}
y^{(2)} = \boldsymbol{\omega}_*^\top\mathbf{x} + \alpha(\theta_o - \boldsymbol{\omega}^\top\mathbf{V}\mathbf{x}) + \eta.
\end{equation}

\subsection{Social Welfare Comparison}

\paragraph{False Negatives and False Positives.}
When agents manipulate under a misaligned threshold (private regime), they may either under- or over-adjust, resulting in:
\begin{itemize}
    \item \textbf{False Negatives (FN):} Qualified agents fail to meet their true acceptance threshold.
    \begin{align*}
    \Delta_{\text{FN}}^A &= N_A \cdot \Pr\left(y^{(2)} \geq y_{\text{th}}, \theta_o < \theta_a\right), \\
    \Delta_{\text{FN}}^B &= N_B \cdot \Pr\left(y^{(2)} \geq y_{\text{th}}, \theta_o > \theta_b\right),
    \end{align*}
    where $N_A$ and $N_B$ are the sizes of each group.
    \item \textbf{False Positives (FP):} Unqualified agents are (incorrectly) accepted due to over- or under-manipulation.
    \begin{align*}
    \Delta_{\text{FP}}^A &= N_A \left[\Pr(y^{(2)} < y_{\text{th}}, \theta_o \leq \theta_a) - \Pr(y^{(1)} < y_{\text{th}})\right], \\
    \Delta_{\text{FP}}^B &= N_B \left[\Pr(y^{(2)} < y_{\text{th}}, \theta_o \geq \theta_b) - \Pr(y^{(1)} < y_{\text{th}})\right].
    \end{align*}
\end{itemize}

\paragraph{Welfare Gap.}
Combining the above, the difference in expected social welfare between public and private regimes is:
\begin{equation}
\mathbb{E}[W_{\text{public}}] - \mathbb{E}[W_{\text{private}}] = \Delta_{\text{FN}}^A + \Delta_{\text{FN}}^B + \Delta_{\text{FP}}^A + \Delta_{\text{FP}}^B \geq 0,
\end{equation}
where the inequality follows because every misalignment of manipulation under private constraints either reduces acceptance of truly qualified agents or increases acceptance of unqualified agents (or both).

\subsection{Quantitative Analysis}

We assume that agents in group $g \in \{A, B\}$ have initial features $x_i \sim \mathcal{N}(\mu_g, \Sigma_g)$, where $\mu_g \in \mathbb{R}^d$ is the group mean and $\Sigma_g \in \mathbb{R}^{d \times d}$ is the covariance matrix.

\noindent
\textbf{Scoring and True Qualification:}
\begin{itemize}
    \item \emph{Scoring function:} $s_i = \omega^\top V x'_i = \omega^\top V (x_i + M a_i)$.
    \item \emph{True qualification:} $y_i = (\omega^*)^\top (x_i + M a_i) + \eta_i$, where $\eta_i \sim \mathcal{N}(0, \sigma^2)$.
\end{itemize}

\noindent
\textbf{Thresholds:}
\begin{itemize}
    \item \emph{Public regime (Scenario 1):} $\theta_A > \theta_0 > \theta_B$.
    \item \emph{Private regime (Scenario 2):} Agents mistakenly apply $\theta_0$, but the system actually uses $\theta_A$ and $\theta_B$ for $A$ and $B$ respectively.
\end{itemize}

\paragraph{Scenario 1: Social Welfare under Public Fairness Constraints ($\mathbb{E}[W_1]$)}

Agents optimally adjust $a_i$ such that $s_i = \theta_g$. Therefore,
\begin{equation}
a_i = \frac{\theta_g - \omega^\top V x_i}{\omega^\top V M}
\end{equation}
The true qualification is then
\begin{equation}
y_i = (\omega^*)^\top x_i + \frac{(\omega^*)^\top M}{\omega^\top V M} (\theta_g - \omega^\top V x_i) + \eta_i
\end{equation}
Let $\omega^* M \neq 0$. Then, $y_i$ is Gaussian with mean and variance:
\begin{align*}
\mu_{y_g} &= \mathbb{E}[y_i] = (\omega^*)^\top \mu_g + \frac{(\omega^*)^\top M}{\omega^\top V M} (\theta_g - \omega^\top V \mu_g) \\
\sigma_{y_g}^2 &= \text{Var}(y_i) \\
&= \left( \omega^* - \frac{(\omega^*)^\top M}{\omega^\top V M} \omega^\top V \right) \Sigma_g \left( \omega^* - \frac{(\omega^*)^\top M}{\omega^\top V M} \omega^\top V \right)^\top
\end{align*}
Thus, the probability that an agent qualifies is:
\begin{equation}
\Pr(y_i \geq y_{\text{thresh}}) = \Phi\left( \frac{\mu_{y_g} - y_{\text{thresh}}}{\sqrt{\sigma_{y_g}^2 + \sigma^2}} \right)
\end{equation}
where $\Phi(\cdot)$ is the standard normal CDF.

\noindent\textbf{Expected Social Welfare:}
\begin{equation}
\mathbb{E}[W_1] = N_A \cdot \Phi\left( \frac{\mu_{y_A} - y_{\text{thresh}}}{\sqrt{\sigma_{y_A}^2 + \sigma^2}} \right) + N_B \cdot \Phi\left( \frac{\mu_{y_B} - y_{\text{thresh}}}{\sqrt{\sigma_{y_B}^2 + \sigma^2}} \right)
\end{equation}

\paragraph{Scenario 2: Social Welfare under Private Fairness Constraints ($\mathbb{E}[W_2]$)}

Agents mistakenly adjust for $\theta_0$:
\begin{equation}
a_i = \frac{\theta_0 - \omega^\top V x_i}{\omega^\top V M}
\end{equation}
and
\begin{equation}
y_i = (\omega^*)^\top x_i + \frac{(\omega^*)^\top M}{\omega^\top V M} (\theta_0 - \omega^\top V x_i) + \eta_i
\end{equation}
However, the actual classification thresholds are $\theta_A$ and $\theta_B$, leading to two types of losses.

\noindent\textbf{False Negative Loss (FN):}
\begin{equation}
\Pr(y_i \geq y_{\text{thresh}}, s_i < \theta_B) = \Phi\left( \frac{\mu_{y_B} - y_{\text{thresh}}}{\sqrt{\sigma_{y_B}^2 + \sigma^2}} \right) \cdot \Phi\left( \frac{\theta_B - \theta_0}{\sigma_s} \right)
\end{equation}
where $\sigma_s = \sqrt{ \omega^\top V \Sigma_g V^\top \omega }$.

\noindent\textbf{False Positive Loss (FP):}
\begin{equation}
\Pr(y_i < y_{\text{thresh}}, s_i \geq \theta_g) = \Phi\left( \frac{y_{\text{thresh}} - \mu_{y_g}}{\sqrt{\sigma_{y_g}^2 + \sigma^2}} \right) \cdot \left[1 - \Phi\left( \frac{\theta_g - \theta_0}{\sigma_s} \right)\right]
\end{equation}

\noindent\textbf{Expected Social Welfare:}
\begin{equation}
\mathbb{E}[W_2] = \sum_{g \in \{A, B\}} N_g \cdot \Phi\left( \frac{\mu_{y_g} - y_{\text{thresh}}}{\sqrt{\sigma_{y_g}^2 + \sigma^2}} \right) \cdot \left[1 - \Phi\left( \frac{\theta_g - \theta_0}{\sigma_s} \right)\right]
\end{equation}

\paragraph{Welfare Difference: $\mathbb{E}[W_1] - \mathbb{E}[W_2]$}

\begin{equation}
\mathbb{E}[W_1] - \mathbb{E}[W_2] = \sum_{g \in \{A, B\}} N_g \cdot \Phi\left( \frac{\mu_{y_g} - y_{\text{thresh}}}{\sqrt{\sigma_{y_g}^2 + \sigma^2}} \right) \cdot \Phi\left( \frac{\theta_g - \theta_0}{\sigma_s} \right)
\end{equation}
Given $\theta_A > \theta_0 > \theta_B$, all terms are positive, so $\mathbb{E}[W_1] - \mathbb{E}[W_2] > 0$.

\textbf{Conclusion:} \textit{Public disclosure of group fairness constraints aligns agents' optimization with actual classification criteria and improves overall social welfare.}

\section{B. Proof of Proposition 1}

In this section, we formally analyze how the visibility of fairness constraints affects the manipulation costs incurred by agents. Specifically, we show that concealing group-specific thresholds leads to strictly higher total manipulation costs compared to disclosing them, under a broad class of cost functions.

Let $c(\cdot, \cdot)$ denote a manipulation cost function (e.g., a distance or norm in feature or score space) that satisfies the triangle inequality and is monotonic in its arguments. For an agent with pre-manipulation score $\theta_x$, let $\theta_o$ be the baseline threshold, and let $\theta_i \in \{\theta_b, \theta_a\}$ denote the group-specific threshold.

\paragraph{Public Thresholds.} Agents are aware of their group membership and adjust features to precisely meet the group-specific threshold:
\begin{equation}
\mathrm{cost}_{\text{pub}} = c(\theta_x, \theta_i)
\end{equation}

\paragraph{Private Thresholds.} Agents, unaware of their true group threshold, initially aim for $\theta_o$. Upon discovering the actual threshold, they must re-adjust:
\begin{equation}
\mathrm{cost}_{\text{hid}} = c(\theta_x, \theta_o) + c(\theta_o, \theta_i)
\end{equation}

By the triangle inequality, $c(\theta_x, \theta_i) \leq c(\theta_x, \theta_o) + c(\theta_o, \theta_i)$, with strict inequality in general. Therefore, concealing group-specific thresholds systematically increases the burden on all agents.

\begin{table}[h]
\centering
\begin{tabular}{llcc}
\toprule
\textbf{Distance Metric} & \textbf{Definition} & \textbf{Triangle Ineq.} & $\mathrm{cost}_{\text{pub}} < \mathrm{cost}_{\text{hid}}$? \\
\midrule
$L_p$ norm & $\|x - x'\|_p$, $p \ge 1$ & Yes & Yes \\
Mahalanobis & $\sqrt{(x - x')^\top\Sigma^{-1}(x - x')}$ & Yes & Yes \\
Any norm/seminorm & Standard norm properties & Yes & Yes \\
Graph distance & Shortest path length & Yes & Yes \\
\bottomrule
\end{tabular}
\caption{Common manipulation cost functions and their properties.}
\label{tab:manipulation-costs}
\end{table}

\section{C. Proof of Theorem 2}
\label{app:belief-convergence}

Let $\mathcal{M} = \{m_1, \ldots, m_M\}$ be the set of candidate mechanisms. The true mechanism is $m^* \in \mathcal{M}$. The agent maintains a belief vector $b_t = (b_t^{(m)})_{m \in \mathcal{M}}$ at round $t$, with all $b_0^{(m)} > 0$.

At each round $t$, the Bayesian update is:
\begin{equation}
b_{t+1}^{(m)} = \frac{b_t^{(m)} \cdot P_{f^{(m)}}(y_t|x'_t)}{\sum_{m' \in \mathcal{M}} b_t^{(m')} \cdot P_{f^{(m')}}(y_t|x'_t)}
\end{equation}

By recursion, after $t$ rounds:
\begin{equation}
b_t^{(m)} = \frac{b_0^{(m)} \prod_{s=1}^t P_{f^{(m)}}(y_s|x'_s)}{\sum_{m' \in \mathcal{M}} b_0^{(m')} \prod_{s=1}^t P_{f^{(m')}}(y_s|x'_s)}.
\end{equation}

For any incorrect $m \neq m^*$, the ratio $R_t^{(m)} = b_t^{(m)}/b_t^{(m^*)}$ satisfies:
\begin{equation}
\log R_t^{(m)} = \log \frac{b_0^{(m)}}{b_0^{(m^*)}} + \sum_{s=1}^t \log \frac{P_{f^{(m)}}(y_s|x'_s)}{P_{f^{(m^*)}}(y_s|x'_s)}.
\end{equation}

By the Strong Law of Large Numbers, since $y_s \sim P_{f^{(m^*)}}(\cdot|x'_s)$:
\begin{equation}
\frac{1}{t} \sum_{s=1}^t \log \frac{P_{f^{(m)}}(y_s|x'_s)}{P_{f^{(m^*)}}(y_s|x'_s)} \to -D_{KL}(P_{f^{(m^*)}} \Vert P_{f^{(m)}}) < 0.
\end{equation}

Let $\delta = \min_{m \neq m^*} D_{KL}(P_{f^{(m^*)}} \Vert P_{f^{(m)}}) > 0$. Then $R_t^{(m)} \leq \frac{b_0^{(m)}}{b_0^{(m^*)}} e^{-\delta t + o(t)} \to 0$ exponentially. Since beliefs are normalized, $b_t^{(m^*)} = \frac{1}{1 + \sum_{m \neq m^*} R_t^{(m)}} \to 1$.

\begin{remark}
    As long as the true mechanism is statistically distinguishable from all alternatives (positive KL divergence), Bayesian updating guarantees convergence to the true mechanism at an exponential rate.
\end{remark}

\section{D. Proof of Theorem 3}
\label{appD}

\subsection{Problem Setting}

\begin{itemize}
    \item Groups: $g \in \{A, B\}$, where $A$ is advantaged and $B$ is disadvantaged.
    \item Thresholds: $\theta_A > \theta_B$ are group-specific.
    \item Scoring: $s(\bx) = \bomega^\top \bx$.
    \item Manipulation cost: $c(\bx, \bx') = \frac{1}{2} \| \bx' - \bx \|_{\bSigma^{-1}}^2$.
\end{itemize}

Under public disclosure, each group agent solves optimally for $\theta_g$, giving acceptance probability $\mathbb{E}[s_g^{\text{Public}}] = \Phi\left( \frac{\theta_g - \mu_g}{\sigma_g} \right)$.

In the PFA setting, agents maintain beliefs over $\Theta = \{\theta_1, \ldots, \theta_M\}$ and update via:
\begin{equation}
b_{t+1}^{(m)} \propto b_t^{(m)} \cdot \exp\left( -\frac{(s(\bx_t') - \theta_m)^2}{2\sigma^2} \right).
\end{equation}

This leads to more conservative early behavior, with the risk of reversal decaying as:
\begin{equation}
\Pr(\Delta_{\text{Fairness}} < 0) \leq \exp\left( -\frac{T(\theta_A - \theta_B)^2}{8\sigma^2} \right) + \frac{C}{\eta T},
\end{equation}
which decays exponentially in $T$, unlike the public case where the risk remains $\Omega(1)$.

\section{E. Proof of Theorem 4}
\label{appE}

Social welfare is defined as:
\begin{equation}
W = \sum_{g \in \{A,B\}} \sum_{i \in g} \mathbb{I}(y_i \geq y_{\text{th}}) \cdot \mathbb{I}(s(x_i') \geq \theta_g) - \lambda \sum_{g,i} c(x_i, x_i').
\end{equation}

Under private fairness, agents optimize for $\theta_o$ instead of the true $\theta_g$, leading to false negatives for group $B$ and false positives for group $A$, plus wasted manipulation cost. Under PFA, beliefs converge to the true threshold (Theorem 2), so both groups approach optimal manipulation as $t \to \infty$. The total welfare gap satisfies:
\begin{equation}
W_{\text{PFA}} - W_{\text{private}} \geq \sum_{t=1}^\infty \gamma^{t-1} \Delta W_t + \lambda \Delta \text{Cost} > 0,
\end{equation}
where all terms are nonneg and strictly positive due to early-phase risk hedging and late-phase efficient adaptation.

\begin{figure*}[t]
    \centering
    \begin{subfigure}{0.237\textwidth}
    \centering
    \includegraphics[width=\linewidth]{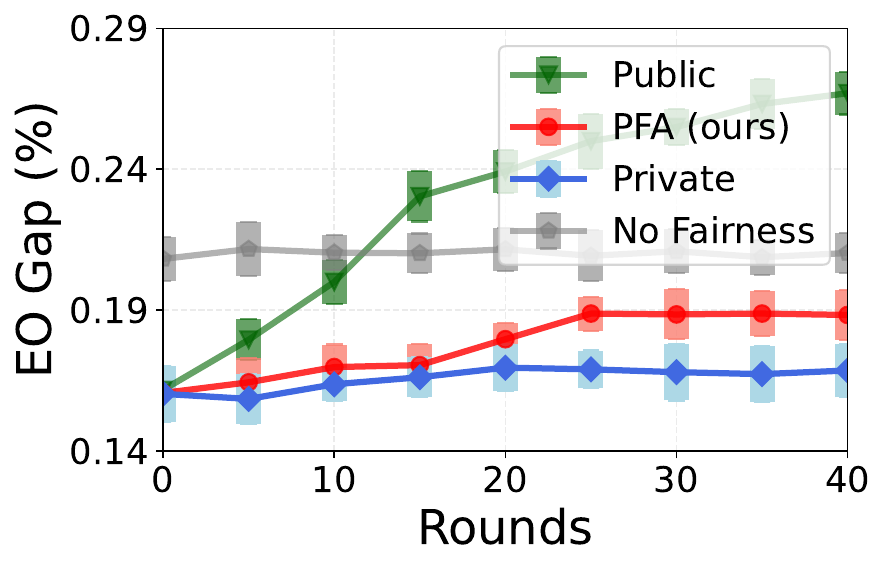}
    \caption{Adult Dataset}
    \label{fig4a}
\end{subfigure}
    \begin{subfigure}{0.237\textwidth}
    \centering
    \includegraphics[width=\linewidth]{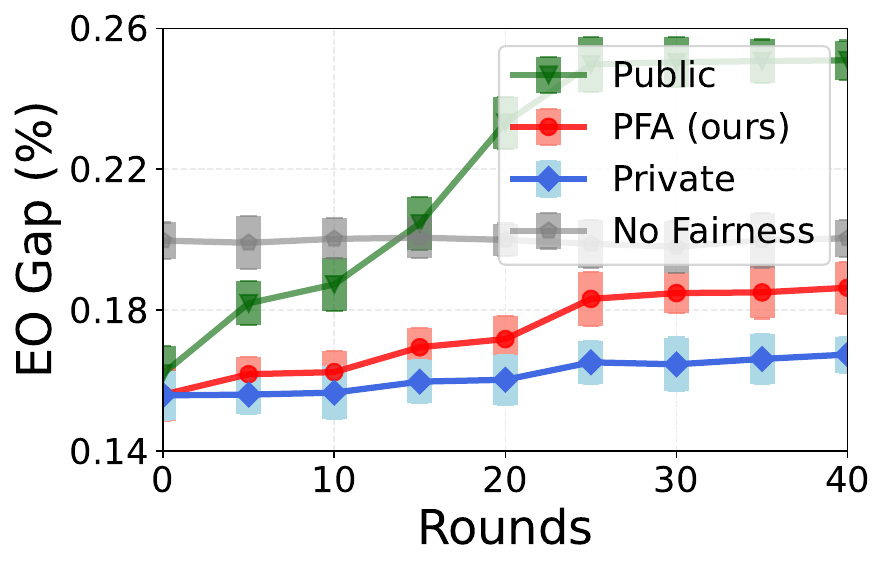}
    \caption{Credit Dataset}
    \label{fig4b}
\end{subfigure}
    \begin{subfigure}{0.237\textwidth}
    \centering
    \includegraphics[width=\linewidth]{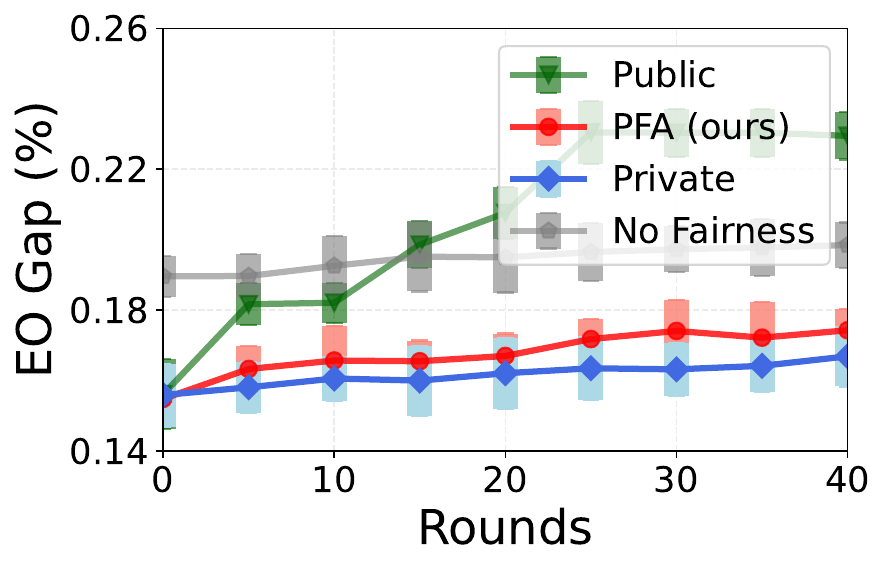}
    \caption{Diabetes Dataset}
    \label{fig4c}
\end{subfigure}
    \begin{subfigure}{0.237\textwidth}
    \centering
    \includegraphics[width=\linewidth]{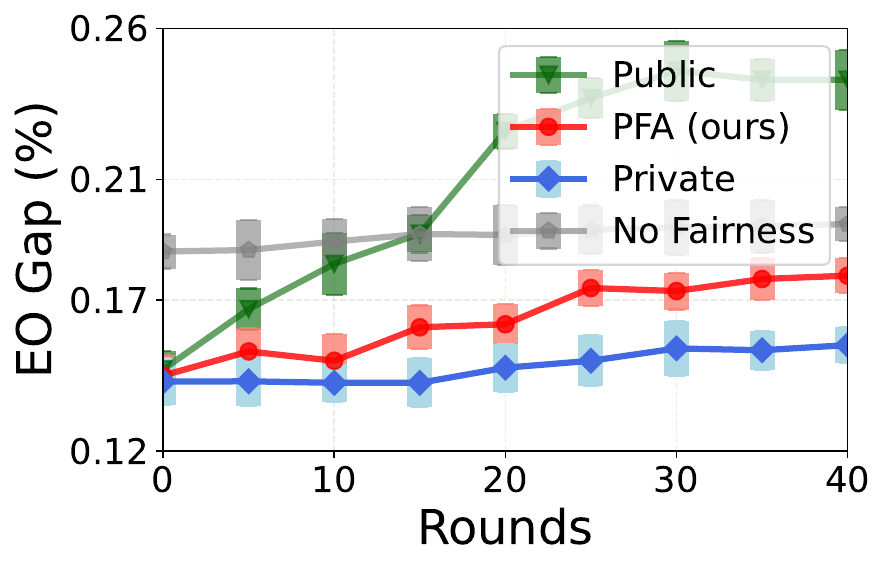}
    \caption{Synthetic Dataset}
    \label{fig4d}
\end{subfigure}
    \caption{Performance of equal odds gap on different real-world and synthetic datasets.}
    \label{fig4}
\end{figure*}

\section{F. More Experimental Results}

\subsection{Additional Metric}
In addition to the fairness metrics discussed in the main text, we also evaluate our models using the \textit{Equalized Odds (EO) gap}, which measures the discrepancy in both true positive rates (TPR) and false positive rates (FPR) across groups:
\begin{equation}
\mathrm{EO\ gap} = \max_{g, g' \in \mathcal{G}} \left| \mathrm{TPR}_g - \mathrm{TPR}_{g'} \right| + \max_{g, g' \in \mathcal{G}} \left| \mathrm{FPR}_g - \mathrm{FPR}_{g'} \right|.
\end{equation}

\subsection{Additional Results and Analysis}

Figure~\ref{fig4} shows that our PFA method consistently achieves the lowest Equalized Odds (EO) gap across all datasets, significantly outperforming both Public and Private baselines. In contrast, the Public mechanism results in the highest EO gap, indicating greater fairness disparities when thresholds are fully disclosed. These results confirm that PFA more effectively balances TPR and FPR between groups, robustly mitigating fairness reversal risks in both real-world and synthetic settings.

Table~\ref{tab3} demonstrates that the EO gap of our PFA method remains stable across a range of learning rates~($\eta$), with only slight increases as $\eta$ grows. This confirms the robustness of PFA's fairness performance to different learning rate choices throughout the learning process.

\begin{table}[h]
\centering
\caption{Performance of equal odds gap for different learning rates $\eta$ and rounds $T$.}
\label{tab3}
\renewcommand{\arraystretch}{1.05}
\begin{tabular}{lcccccc}
\toprule
 \textbf{\textit{EO Gap}}(\%) & $t=5$ & $10$ & $15$ & $20$ & $30$ & $40$ \\
\midrule
 \textit{$\eta=$ 0.01} & 0.170 & 0.173 & 0.177 & 0.181 & 0.190 & 0.194 \\
 \textit{$\eta=$ 0.05} & 0.173 & 0.179 & 0.183 & 0.191 & 0.198 & 0.201 \\
 \textit{$\eta=$ 0.1}  & 0.182 & 0.187 & 0.192 & 0.198 & 0.202 & 0.205 \\
 \textit{$\eta=$ 0.5}  & 0.190 & 0.196 & 0.202 & 0.208 & 0.214 & 0.217 \\
\bottomrule
\end{tabular}
\end{table}

\end{document}